\documentclass{article} 
\usepackage{iclr2021_conference,times}

\usepackage{amsmath,amsfonts,bm}

\def\eqref#1{equation~\ref{#1}}

\def\1{\bm{1}}

\DeclareMathAlphabet{\mathsfit}{\encodingdefault}{\sfdefault}{m}{sl}
\SetMathAlphabet{\mathsfit}{bold}{\encodingdefault}{\sfdefault}{bx}{n}

\usepackage[utf8]{inputenc} 
\usepackage[T1]{fontenc}    
\usepackage{hyperref}       
\hypersetup{colorlinks,allcolors=black}
\usepackage{url}            
\usepackage{booktabs}       
\usepackage{amsmath}
\usepackage{amssymb}
\usepackage{amsfonts}       
\usepackage{commath}
\usepackage{nicefrac}       
\usepackage{bm}
\usepackage{microtype}      
\usepackage{float}
\usepackage{tabularx}
\usepackage{xcolor}
\usepackage{graphicx}
\usepackage{subcaption}
\usepackage{lipsum}
\usepackage{adjustbox}
\usepackage{wrapfig}

\usepackage[bottom]{footmisc}

\definecolor{spectral1}{HTML}{9E0142}
\definecolor{spectral2}{HTML}{D53E4F}
\definecolor{spectral3}{HTML}{F46D43}
\definecolor{spectral4}{HTML}{FDAE61}
\definecolor{spectral5}{HTML}{606669} 
\definecolor{spectral6}{HTML}{FFFFBF}
\definecolor{spectral7}{HTML}{E6F598}
\definecolor{spectral8}{HTML}{ABDDA4}
\definecolor{spectral9}{HTML}{66C2A5}
\definecolor{spectral10}{HTML}{3288BD}
\definecolor{spectral11}{HTML}{5E4FA2}

\usepackage{pgfplots}
\pgfplotsset{width=10cm,compat=1.9}
\usetikzlibrary{shapes.geometric,decorations.pathreplacing,matrix,pgfplots.groupplots,pgfplots.external,}
\usepgfplotslibrary{fillbetween}
\usepackage{pgfplotstable}
\usepackage{filecontents}

\hypersetup{
  pdfinfo={
    Title={Gradient Origin Networks},
    Author={Sam Bond-Taylor, Chris G. Willcocks},
    Subject={Machine Learning, Pattern Recognition},
    Keywords={Generative Models, Implicit, Representation Networks, Autoencoders, Neural Networks}
  }
}

\title{Gradient Origin Networks}

\author{Sam Bond-Taylor$^*$ \& Chris G. Willcocks\thanks{Authors contributed equally.} \\
Department of Computer Science\\
Durham University\\
\texttt{\{samuel.e.bond-taylor,christopher.g.willcocks\}@durham.ac.uk} \\
}

\DeclareMathOperator{\reals}{\mathbb{R}}

\iclrfinalcopy 
\begin{document}

\maketitle

\begin{abstract}
  This paper proposes a new type of generative model that is able to quickly learn a latent representation without an encoder. This is achieved using empirical Bayes to calculate the expectation of the posterior, which is implemented by initialising a latent vector with zeros, then using the gradient of the log-likelihood of the data with respect to this zero vector as new latent points. The approach has similar characteristics to autoencoders, but with a simpler architecture, and is demonstrated in a variational autoencoder equivalent that permits sampling. This also allows implicit representation networks to learn a space of implicit functions without requiring a hypernetwork, retaining their representation advantages across datasets. The experiments show that the proposed method converges faster, with significantly lower reconstruction error than autoencoders, while requiring half the parameters.
\end{abstract}

\section{Introduction}
Observable data in nature has some parameters which are known, such as local coordinates, but also some unknown parameters such as how the data is related to other examples. Generative models, which learn a distribution over observables, are central to our understanding of patterns in nature and allow for efficient query of new unseen examples. Recently, deep generative models have received interest due to their ability to capture a broad set of features when modelling data distributions. As such, they offer direct applications such as synthesising high fidelity images \citep{karras2020analyzing}, super-resolution \citep{dai2019second}, speech synthesis \citep{li2019neural}, and drug discovery \citep{segler2018planning}, as well as benefits for downstream tasks like semi-supervised learning \citep{chen2020generative}.

\begin{figure}[b!]
\vspace*{-0.8em}
\centering
\begin{subfigure}{0.24\textwidth}
    \centering
    \begin{tikzpicture}[shorten >=1pt,->]
        \tikzstyle{cir}=[circle,fill=black!30,minimum size=17pt,inner sep=0pt]
        \tikzstyle{dia}=[diamond,fill=black!30,minimum size=19pt,inner sep=0pt]
        \tikzstyle{box}=[rounded corners=6pt,fill=black!25,minimum size=17pt,inner xsep=4pt, inner ysep=0pt]
        \tikzstyle{sqr}=[fill=black!25,minimum size=17pt, inner ysep=0pt, inner xsep=4pt]
        
        \node[cir,fill=black!12]  (VAE-x)    at (0,1) {$\mathbf{x}$};
        \node[dia]                (VAE-mu)   at (1,1.75) {$\boldsymbol\mu$};
        \node[dia]                (VAE-sig)  at (1,1) {$\boldsymbol\sigma$};
        \node[cir,fill=black!12]  (VAE-eps)  at (1,0.25) {$\boldsymbol\epsilon$};
        \node[dia,fill=black!12]  (VAE-z)    at (2,1) {$\mathbf{z}$};
        \node[cir]                (VAE-xh)   at (3,1) {$\mathbf{\hat{x}}$};
        
        \draw(VAE-x) -- (VAE-mu);
        \draw(VAE-x) -- (VAE-sig);
        \draw(VAE-mu) -- (VAE-z);
        \draw(VAE-sig) -- (VAE-z);
        \draw(VAE-eps) -- (VAE-z);
        \draw(VAE-z) -- (VAE-xh);
        
        \node[align=center] at (0.45,0.7) {$E$};
        \node[align=center] at (2.45,1.3) {$D$};
        \node[align=center] at (2.3,0.45) {$\boldsymbol\mu + \boldsymbol\sigma \odot  \boldsymbol\epsilon $};
    \end{tikzpicture}
    \caption{VAE}
    \label{fig:vae}
\end{subfigure}
\hspace{\fill}
\begin{subfigure}{0.24\textwidth}
    \centering
    \vspace*{1.3em}
    \begin{tikzpicture}[shorten >=1pt,->]
        \tikzstyle{cir}=[circle,fill=black!30,minimum size=17pt,inner sep=0pt]
        \tikzstyle{dia}=[diamond,fill=black!30,minimum size=19pt,inner sep=0pt]
        \tikzstyle{box}=[rounded corners=6pt,fill=black!25,minimum size=17pt,inner xsep=4pt, inner ysep=0pt]
        \tikzstyle{sqr}=[fill=black!25,minimum size=17pt, inner ysep=0pt, inner xsep=4pt]

        \draw[rounded corners=0pt,->] (0,1.5) -- (0,1.0) -- (0.7,1.0);
        
        \node[dia,fill=black!12]                (GON-0)    at (0,1.5) {$\mathbf{0}$};
        \node[cir]                (GON-2)    at (1,1.0) {$\mathbf{\hat{x}}$};
        
        \draw[densely dashed] (GON-2.north) to [out=90,in=0] (GON-0.east);
        \node[align=center] at (0.42,1.2) {$F$};
        
    \end{tikzpicture}
    \vspace*{1.5em}
    \caption{GON}
    \label{fig:gon}
\end{subfigure}
\hspace{\fill}
\begin{subfigure}{0.24\textwidth}
    \centering
    \vspace*{1.3em}
    \begin{tikzpicture}[shorten >=1pt,->]
        \tikzstyle{cir}=[circle,fill=black!30,minimum size=17pt,inner sep=0pt]
        \tikzstyle{dia}=[diamond,fill=black!30,minimum size=19pt,inner sep=0pt]
        \tikzstyle{box}=[rounded corners=6pt,fill=black!25,minimum size=17pt,inner xsep=4pt, inner ysep=0pt]
        \tikzstyle{sqr}=[fill=black!25,minimum size=17pt, inner ysep=0pt, inner xsep=4pt]
        
        \draw (0,1) -- (0,0);
        
        \node[dia,fill=black!12]                (GON-0)    at (0,1) {$\mathbf{0}$};
        \node[cir,fill=black!12]  (GON-1)    at (0,0) {$\mathbf{c}$};
        \node[cir]                (GON-2)    at (1,0.5) {$\mathbf{\hat{x}}$};
        \draw (0,0.5) -> (GON-2);
        \draw[densely dashed] (GON-2.north) to [out=90,in=0] (GON-0.east);
        \node[align=center] at (0.42,0.7) {$F$};
        
    \end{tikzpicture}
    \caption{Implicit GON}
    \label{fig:implicit-gon}
\end{subfigure}
\hspace{\fill}
\begin{subfigure}{0.24\textwidth}
    \centering
    \begin{tikzpicture}[shorten >=1pt,->]
        \tikzstyle{cir}=[circle,fill=black!30,minimum size=17pt,inner sep=0pt]
        \tikzstyle{dia}=[diamond,fill=black!30,minimum size=19pt,inner sep=0pt]
        \tikzstyle{box}=[rounded corners=6pt,fill=black!25,minimum size=17pt,inner xsep=4pt, inner ysep=0pt]
        \tikzstyle{sqr}=[fill=black!25,minimum size=17pt, inner ysep=0pt, inner xsep=4pt]
        
        \node[dia,fill=black!12]  (VAE-0)    at (2,1.75) {$\mathbf{0}$};
        \node[dia]                (VAE-mu)   at (1,1.75) {$\boldsymbol\mu$};
        \node[dia]                (VAE-sig)  at (1,1) {$\boldsymbol\sigma$};
        \node[cir,fill=black!12]  (VAE-eps)  at (1,0.25) {$\boldsymbol\epsilon$};
        \node[dia,fill=black!12]  (VAE-z)    at (2,1) {$\mathbf{z}$};
        \node[cir]                (VAE-xh)   at (3,1) {$\mathbf{\hat{x}}$};
        
        \draw(VAE-0) -- (VAE-mu);
        \draw(VAE-0) -- (VAE-sig);
        \draw(VAE-mu) -- (VAE-z);
        \draw(VAE-sig) -- (VAE-z);
        \draw(VAE-eps) -- (VAE-z);
        \draw(VAE-z) -- (VAE-xh);
        
        \draw[densely dashed] (VAE-xh.north) to [out=90,in=0] (VAE-0.east);
        
        \node[align=center] at (2.45,1.3) {$F$};
        \node[align=center] at (2.3,0.45) {$\boldsymbol\mu + \boldsymbol\sigma \odot  \boldsymbol\epsilon $};
    \end{tikzpicture}
    \caption{Variational GON}
    \label{fig:vae-gon}
\end{subfigure}
\caption{Gradient Origin Networks (GONs; b) use gradients (dashed lines) as encodings thus only a single network $F$ is required, which can be an implicit representation network (c). Unlike VAEs (a) which use two networks, $E$ and $D$, variational GONs (d) permit sampling with only one network.}
\label{fig:network-diagrams}
\end{figure}
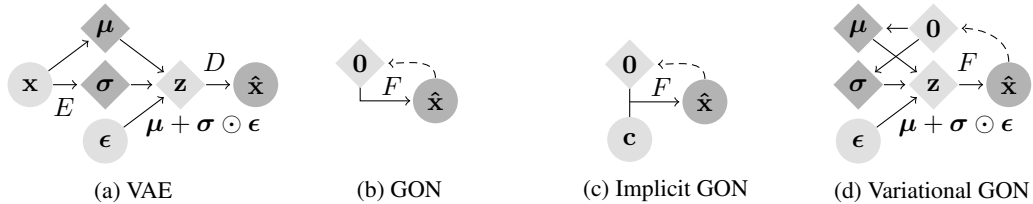

A number of methods have been proposed such as Variational Autoencoders (VAEs, Figure~\ref{fig:vae}), which learn to encode the data to a latent space that follows a normal distribution permitting sampling \citep{Kingma2014AutoEncodingVariationalBayes}. Generative Adversarial Networks (GANs) have two competing networks, one which generates data and another which discriminates from implausible results \citep{Goodfellow2014GenerativeAdversarialNets}. Variational approaches that approximate the posterior using gradient descent \citep{lipton2017precise} and short run MCMC \citep{Nijkamp2020LearningMultilayerLatent} respectively have been proposed, but to obtain a latent vector for a sample, they require iterative gradient updates. Autoregressive Models \citep{VanDenOord2016PixelRecurrentNeural} decompose the data distribution as the product of conditional distributions and Normalizing Flows \citep{Rezende2015VariationalInferenceNormalizing} chain together invertible functions; both methods allow exact likelihood inference. Energy-Based Models (EBMs) map data points to energy values proportional to likelihood thereby permitting sampling through the use of Monte Carlo Markov Chains \citep{Du2019ImplicitGenerationGeneralization}. In general to support encoding, these approaches require separate encoding networks, are limited to invertible functions, or require multiple sampling steps.

Implicit representation learning \citep{Park2019DeepSDFLearningContinuous,Tancik2020FourierFeaturesLet}, where a network is trained on data parameterised continuously rather than in discrete grid form, has seen a surge of interest due to the small number of parameters, speed of convergence, and ability to model fine details. In particular, sinusoidal representation networks (SIRENs) \citep{Sitzmann2020ImplicitNeuralRepresentationsa} achieve impressive results, modelling many signals with high precision, thanks to their use of periodic activations paired with carefully initialised MLPs. So far, however, these models have been limited to modelling single data samples, or use an additional hypernetwork or meta learning \citep{Sitzmann2020MetaSDFMetalearningSigned} to estimate the weights of a simple implicit model, adding significant complexity.

This paper proposes Gradient Origin Networks (GONs), a new type of generative model (Figure~\ref{fig:gon}) that do not require encoders or hypernetworks. This is achieved by initialising latent points at the origin, then using the gradient of the log-likelihood of the data with respect to these points as the latent space. At inference, latent vectors can be obtained in a single step without requiring iteration. GONs are shown to have similar characteristics to convolutional autoencoders and variational autoencoders using approximately half the parameters, and can be applied to implicit representation networks (such as SIRENs) allowing a space of implicit functions to be learned with a simpler overall architecture.

\section{Preliminaries}




We first introduce some background context that will be used to derive our proposed approach.

\subsection{Empirical Bayes}


\newcommand\negpad{-0.501px}

The concept of empirical Bayes \citep{robbins1956empirical,saremi2019neural}, for a random variable $\mathbf{z} \sim p_\mathbf{z}$ and particular observation $\mathbf{z}_0 \sim p_{\mathbf{z}_0}$, provides an estimator of $\mathbf{z}$ expressed purely in terms of $p({\mathbf{z}_0})$ that minimises the expected squared error. This estimator can be written as a conditional mean:
\vspace*{-6px}
\begin{equation}
    \hat{\mathbf{z}}(\mathbf{z}_0) = \int \mathbf{z} p(\mathbf{z}|\mathbf{z}_0) d\mathbf{z} = \int \mathbf{z} \frac{p(\mathbf{z},\mathbf{z}_0)}{p(\mathbf{z}_0)} d\mathbf{z}. \vspace*{\negpad}
    \label{eqn:bayes-estimator}
\end{equation}
Of particular relevance is the case where $\mathbf{z}_0$ is a noisy observation of $\mathbf{z}$ with covariance $\bm{\Sigma}$. In this case $p(\mathbf{z}_0)$ can be represented by marginalising out $\mathbf{z}$:
\vspace*{\negpad}
\begin{equation}
    p(\mathbf{z}_0) = \int \frac{1}{(2\pi)^{d/2}|\det(\bm{\Sigma})|^{1/2}}\exp \Big( -(\mathbf{z}_0-\mathbf{z})^T \bm{\Sigma}^{-1}(\mathbf{z}_0-\mathbf{z})/2 \Big) p(\mathbf{z}) d\mathbf{z}. \vspace*{\negpad}
\end{equation}
Differentiating this with respect to $\mathbf{z}_0$ and multiplying both sides by $\bm{\Sigma}$ gives:
\vspace*{\negpad}
\begin{equation}
    \bm{\Sigma} \nabla_{\mathbf{z}_0} p(\mathbf{z}_0) = \int (\mathbf{z} - \mathbf{z}_0) p(\mathbf{z},\mathbf{z}_0) d\mathbf{z} = \int \mathbf{z} p(\mathbf{z},\mathbf{z}_0) d\mathbf{z} - \mathbf{z}_0p(\mathbf{z}_0). \vspace*{\negpad}
\end{equation}
After dividing through by $p(\mathbf{z}_0)$ and combining with Equation~\ref{eqn:bayes-estimator} we obtain a closed form estimator of $\mathbf{z}$ \citep{miyasawa1961empirical} written in terms of the score function $\nabla \log p(\mathbf{z}_0)$ \citep{hyvarinen2005estimation}:
\vspace*{\negpad}
\begin{equation}
    \hat{\mathbf{z}}(\mathbf{z}_0) = \mathbf{z}_0 + \bm{\Sigma} \nabla_{\mathbf{z}_0} \log p(\mathbf{z}_0). \vspace*{\negpad}
    \label{eqn:gaussian-empirical-bayes}
\end{equation}
This optimal procedure is achieved in what can be interpreted as a single gradient descent step, with no knowledge of the prior $p(\mathbf{z})$. By rearranging Equation~\ref{eqn:gaussian-empirical-bayes}, a definition of $\nabla \log p(\mathbf{z}_0)$ can be derived; this can be used to train models that approximate the score function \citep{song2019generative}.

\subsection{Variational Autoencoders}
Variational Autoencoders (VAEs; \citealt{Kingma2014AutoEncodingVariationalBayes}) are a probabilistic take on standard autoencoders that permit sampling. A latent-based generative model $p_{\boldsymbol{\theta}}(\mathbf{x}|\mathbf{z})$ is defined with a normally distributed prior over the latent variables, $p_{\boldsymbol{\theta}}(\mathbf{z})=\mathcal{N}(\mathbf{z};\mathbf{0},\bm{I}_d)$. $p_{\boldsymbol{\theta}}(\mathbf{x}|\mathbf{z})$ is typically parameterised as a Bernoulli, Gaussian, multinomial distribution, or mixture of logits. In this case, the true posterior $p_{\boldsymbol{\theta}}(\mathbf{z}|\mathbf{x})$ is intractable, so a secondary encoding network $q_{\boldsymbol{\phi}}(\mathbf{z}|\mathbf{x})$ is used to approximate the true posterior; the pair of networks thus resembles a traditional autoencoder. This allows VAEs to approximate $p_{\boldsymbol{\theta}}(\mathbf{x})$ by maximising the evidence lower bound (ELBO), defined as:
\vspace*{\negpad}
\begin{equation}
    \log p_{\boldsymbol{\theta}}(\mathbf{x}) \geq \mathcal{L}^{\textup{VAE}} = - D_{\textup{KL}}(\mathcal{N}(q_{\boldsymbol\phi}(\mathbf{z}|\mathbf{x})) || \mathcal{N}(\mathbf{0}, \bm{I}_d)) + \mathbb{E}_{q_{\boldsymbol{\phi}}(\mathbf{z}|\mathbf{x})}[ \log p_{\boldsymbol{\theta}}(\mathbf{x}|\mathbf{z}) ].
\end{equation}
%
To optimise this lower bound with respect to ${\boldsymbol{\theta}}$ and ${\boldsymbol{\phi}}$, gradients must be backpropagated through the stochastic process of generating samples from $\mathbf{z}' \sim q_{\boldsymbol{\phi}}(\mathbf{z}|\mathbf{x})$. This is permitted by reparameterising $\mathbf{z}$ using the differentiable function $\mathbf{z}'= \mu(\mathbf{z}) + \sigma(\mathbf{z}) \odot \boldsymbol{\epsilon}$, where $\boldsymbol{\epsilon} \sim \mathcal{N}(\mathbf{0}, \bm{I}_d)$ and $\mu(\mathbf{z})$ and $\sigma(\mathbf{z})^2$ are the mean and variance respectively of a multivariate Gaussian distribution with diagonal covariance.



\section{Method}

Consider some dataset $\mathbf{x} \sim p_d$ of continuous or discrete signals $\mathbf{x} \in \reals^m$, it is typical to assume that the data can be represented by low dimensional latent variables $\mathbf{z} \in \reals^k$, which can be used by a generative neural network to reconstruct the data. These variables are often estimated through the use of a secondary encoding network that is trained concurrently with the generative network. An encoding network adds additional complexity (and parameters) to the model, it can be difficult to balance capacities of the two networks, and for complex hierarchical generative models designing a suitable architecture can be difficult. This has led some to instead approximate latent variables by performing gradient descent on the generative network \citep{bojanowski2018optimizing,Nijkamp2020LearningMultilayerLatent}. While this addresses the aforementioned problems, it significantly increases the run time of the inference process, introduces additional hyperparameters to tune, and convergence is not guaranteed. 

\subsection{Gradient Origin Networks \label{sec:gons}}

We propose a generative model that consists only of a decoding network, using empirical Bayes to approximate the posterior in a single step. That is, for some data point $\mathbf{x}$ and latent variable $\mathbf{z} \sim p_\mathbf{z}$, we wish to find an approximation of $p(\mathbf{z}|\mathbf{x})$. Given some noisy observation $\mathbf{z}_0=\mathbf{z}+\mathcal{N}(\mathbf{0},\bm{I}_d)$ of $\mathbf{z}$ then empirical Bayes can be applied to approximate $\mathbf{z}$. Specifically, since we wish to approximate $\mathbf{z}$ conditioned on $\mathbf{x}$, we instead calculate $\hat{\mathbf{z}}_\mathbf{x}$, the least squares estimate of $p(\mathbf{z}|\mathbf{x})$ (proof in Appendix~\ref{apx:empirical-bayes-proof}):
\begin{equation}
    \hat{\mathbf{z}}_\mathbf{x}(\mathbf{z}_0) = \mathbf{z}_0 + \nabla_{\mathbf{z}_0} \log p(\mathbf{z}_0|\mathbf{x}).
    \label{eqn:gon-bayes-simple}
\end{equation}
Using Bayes' rule, $\log p(\mathbf{z}_0|\mathbf{x})$ can be written as $\log p(\mathbf{z}_0|\mathbf{x}) = \log p(\mathbf{x}|\mathbf{z}_0) + \log p(\mathbf{z}_0) - \log p(\mathbf{x})$. Since $\log p(\mathbf{x})$ is a normalising constant that does not affect the gradient, we can rewrite Equation~\ref{eqn:gon-bayes-simple} in terms only of the decoding network and $p(\mathbf{z}_0)$:
\begin{equation}
    \hat{\mathbf{z}}_\mathbf{x}(\mathbf{z}_0) = \mathbf{z}_0 + \nabla_{\mathbf{z}_0} \Big( \log p(\mathbf{x}|\mathbf{z}_0) + \log p(\mathbf{z}_0) \Big).
    \label{eqn:gon-bayes-equation}
\end{equation}
It still remains, however, how to construct a noisy estimate of $\mathbf{z}_0$ with no knowledge of $\mathbf{z}$. If we assume $\mathbf{z}$ follows a known distribution, then it is possible to develop reasonable estimates. For instance, if we assume $p(\mathbf{z})=\mathcal{N}(\mathbf{z};\mathbf{0},\bm{I}_d)$ then we could sample from $p(\mathbf{z}_0)=\mathcal{N}(\mathbf{z}_0;\mathbf{0},2\bm{I}_d)$ however this could be far from the true distribution of $p(\mathbf{z}_0|\mathbf{z})=\mathcal{N}(\mathbf{z}_0;\mathbf{z},\bm{I}_d)$. Instead we propose initialising $\mathbf{z}_0$ at the origin since this is the distribution's mean. Initialising at a constant position decreases the input variation and thus simplifies the optimisation procedure. Naturally, how $p(\mathbf{x}|\mathbf{z})$ is modelled affects $\hat{\mathbf{z}}_{\mathbf{x}}$. While mean-field models result in $\hat{\mathbf{z}}_{\mathbf{x}}$ that are linear functions of $\mathbf{x}$, conditional autoregressive models, for instance, result in non-linear $\hat{\mathbf{z}}_{\mathbf{x}}$; multiple gradient steps also induce non-linearity, however, we show that a single step works well on high dimensional data suggesting that linear encoders, which normally do not scale to high dimensional data are effective in this case.





\subsection{Autoencoding with GONs}




Before exploring GONs as generative models, we discuss the case where the prior $p(\mathbf{z})$ is unknown; such a model is referred to as an autoencoder. As such, the distribution $p(\mathbf{z}_0|\mathbf{z})$ is also unknown thus it is again unclear how we can construct a noisy estimate of $\mathbf{z}$. By training a model end-to-end where $\mathbf{z}_0$ is chosen as the origin, however, a prior is implicitly learned over $\mathbf{z}$ such that it is reachable from $\mathbf{z}_0$. Although $p(\mathbf{z})$ is unknown, we do not wish to impose a prior on $\mathbf{z}_0$; the term which enforces this is in Equation~\ref{eqn:gon-bayes-equation} is $\log p(\mathbf{z}_0)$, so we can safely ignore this term and simply maximise the likelihood of the data given $\mathbf{z}_0$. Our estimator of $\mathbf{z}$ can therefore be defined simply as $\hat{\mathbf{z}}_\mathbf{x}(\mathbf{z}_0) = \mathbf{z}_0 + \nabla_{\mathbf{z}_0} \log p(\mathbf{x}|\mathbf{z}_0)$, which can otherwise be interpreted as a single gradient descent step on the conditional log-likelihood of the data. From this estimate, the data can be reconstructed by passing $\hat{\mathbf{z}}_\mathbf{x}$ through the decoder to parameterise $p(\mathbf{x}|\hat{\mathbf{z}}_\mathbf{x})$. This procedure can be viewed more explicitly when using a neural network $F \colon \reals^k \to \reals^m$ to output the mean of $p(\mathbf{x}|\hat{\mathbf{z}}_\mathbf{x})$ parameterised by a normal distribution; in this case the loss function is defined in terms of mean squared error loss $\mathcal{L}^{\text{MSE}}$:
\begin{equation}
    G_{\mathbf{x}} = \mathcal{L}^{\text{MSE}} ( \mathbf{x}, F(-\nabla_{\mathbf{z}_0} \mathcal{L}^{\text{MSE}}(\mathbf{x}, F(\mathbf{z}_0))) ).
    \label{eqn:gon}
\end{equation}
The gradient computation thereby plays a similar role to an encoder, while $F$ can be viewed as a decoder, with the outer loss term determining the overall reconstruction quality. Using a single network to perform both roles has the advantage of simplifying the overall architecture, avoiding the need to balance networks, and avoiding bottlenecks; this is demonstrated in Figure~\ref{fig:gon} which provides a visualisation of the GON process.

\subsection{Variational GONs}





The variational approach can be naturally applied to GONs, allowing sampling in a single step while only requiring the generative network, reducing the parameters necessary. Similar to before, a feedforward neural network $F$ parameterises $p(\mathbf{x}|\mathbf{z})$, while the expectation of $p(\mathbf{z}|\mathbf{x})$ is calculated with empirical Bayes. A normal prior is assumed over $p(\mathbf{z})$ thus Equation~\ref{eqn:gon-bayes-equation} can be written as:
\begin{equation}
    \hat{\mathbf{z}}_\mathbf{x}(\mathbf{z}_0) = \mathbf{z}_0 + \nabla_{\mathbf{z}_0} \Big( \log p(\mathbf{x}|\mathbf{z}_0) + \log \mathcal{N}(\mathbf{z}_0;\mathbf{0},2\bm{I}_d) \Big),
\end{equation}
where $\mathbf{z}_0 = \mathbf{0}$ as discussed in Section~\ref{sec:gons}. While it would be possible to use this estimate directly within a constant-variance VAE \citep{ghosh2019variational}, we opt to incorporate the reparameterisation trick into the generative network as a stochastic layer, to represent the distribution over which $\mathbf{x}$ could be encoded to, using empirical Bayes to estimate $\mathbf{z}$. Similar to the autoencoding approach, we could ignore $\log p(\mathbf{z}_0)$, however we find assuming a normally distributed $\mathbf{z}_0$ implicitly constrains $\mathbf{z}$, aiding the optimisation procedure. Specifically, the forward pass of $F$ is implemented as follows: $\hat{\mathbf{z}}_\mathbf{x}$ (or $\mathbf{z}_0$) is mapped by linear transformations to $\mu(\hat{\mathbf{z}}_\mathbf{x})$ and $\sigma(\hat{\mathbf{z}}_\mathbf{x})$ and the reparameterisation trick is applied, subsequently the further transformations formerly defined as $F$ in the GON formulation are applied, providing parameters for for $p(\mathbf{x}|\hat{\mathbf{z}}_\mathbf{x})$. Training is performed end-to-end, minimising the ELBO:
\begin{equation}
    \mathcal{L}^{\textup{VAE}} = - D_{\textup{KL}}(\mathcal{N}(\mu(\hat{\mathbf{z}}_\mathbf{x}), \sigma(\hat{\mathbf{z}}_\mathbf{x})^2) || \mathcal{N}(\mathbf{0}, \bm{I}_d)) + \log p(\mathbf{x}|\hat{\mathbf{z}}_\mathbf{x}) .
\end{equation}
These steps are shown in Figure~\ref{fig:vae-gon}, which in practice has a simple implementation.


\subsection{Implicit GONs}

In the field of implicit representation networks, the aim is to learn a neural approximation of a function $\Phi$ that satisfies an implicit equation of the form:
\begin{equation}
    R(\mathbf{c}, \Phi, \nabla_\Phi, \nabla^2_\Phi, \dots) = 0, \quad \Phi \colon \mathbf{c} \mapsto \Phi(\mathbf{c}),
\end{equation}
where $R$'s definition is problem dependent but often corresponds to a loss function. Equations with this structure arise in a myriad of fields, namely 3D modelling, image, video, and audio representation \citep{Sitzmann2020ImplicitNeuralRepresentationsa}. In these cases, data samples $\mathbf{x} = \{(\mathbf{c}, \Phi_{\mathbf{x}}(\mathbf{c}))\}$ can be represented in this form in terms of coordinates $\mathbf{c} \in \reals^n$ and the corresponding data at those points $\Phi \colon \reals^n \to \reals^m$. Due to the continuous nature of $\Phi$, data with large spatial dimensions can be represented much more efficiently than approaches using convolutions, for instance. Despite these benefits, representing a distribution of data points using implicit methods is more challenging, leading to the use of hypernetworks which estimate the weights of an implicit network for each data point \citep{Sitzmann2020MetaSDFMetalearningSigned}; this increases the number of parameters and adds significant complexity.

By applying the GON procedure to implicit representation networks, it is possible learn a space of implicit functions without the need for additional networks. We assume there exist latent vectors $\mathbf{z} \in \reals^k$ corresponding to data samples; the concatentation of these latent variables with the data coordinates can therefore geometrically be seen as points on a manifold that describe the full dataset in keeping with the manifold hypothesis \citep{fefferman2016testing}. An implicit Gradient Origin Network can be trained on this space to mimic $\Phi$ using a neural network $F \colon \reals^{n+k} \to \reals^m$, thereby learning a space of implicit functions by modifying Equation~\ref{eqn:gon}:
\begin{equation}
    I_{\mathbf{x}} = \int \mathcal{L} \Big( \Phi_{\mathbf{x}}(\mathbf{c}), F\Big(\mathbf{c} \oplus -\nabla_{\mathbf{z}_0} \int \mathcal{L} \big( \Phi_{\mathbf{x}}(\mathbf{c}), F(\mathbf{c} \oplus \mathbf{z}_0) \big) d\mathbf{c} \Big) \Big) d\mathbf{c},
    \label{eqn:implicit-gon-loss}
\end{equation}
where both integrations are performed over the space of coordinates and $\mathbf{c} \oplus \mathbf{z}$ represents a concatenation (Figure~\ref{fig:implicit-gon}). 
Similar to the non-implicit approach, the computation of latent vectors can be expressed as $\mathbf{z} = -\nabla_{\mathbf{z}_0} \int \mathcal{L} \big( \Phi_{\mathbf{x}}(\mathbf{c}), F(\mathbf{c} \oplus \mathbf{z}_0) \big) d\mathbf{c}$. In particular, we parameterise $F$ with a SIREN \citep{Sitzmann2020ImplicitNeuralRepresentationsa}, finding that it is capable of modelling both high and low frequency components in high dimensional spaces.

\subsection{GON Generalisations}

There are a number of interesting generalisations that make this approach applicable to other tasks. In Equations~\ref{eqn:gon} and \ref{eqn:implicit-gon-loss} we use the same $\mathcal{L}$ in both the inner term and outer term, however, as with variational GONs, it is possible to use different functions; through this, training a GON concurrently as a generative model and classifier is possible, or through some minor modifications to the loss involving the addition of the categorical cross entropy loss function $\mathcal{L}^{\textsc{CCE}}$ to maximise the likelihood of classification, solely as a classifier:
\begin{equation}
    C_{\mathbf{x}} = \mathcal{L}^{\textsc{CCE}}(f( -\nabla_{\mathbf{z}_0} \mathcal{L} ( \mathbf{x}, F(\mathbf{z}_0) ) ), \mathbf{y}),
    \label{eqn:gon-classify}
\end{equation}
where $\mathbf{y}$ are the target labels and $f$ is a single linear layer followed by softmax. Another possibility is modality conversion for translation tasks; in this case the inner reconstruction loss is performed on the source signal and the outer loss on the target signal.

\subsection{Justification}

Beyond empirical Bayes, we provide some additional analysis on why a single gradient step is in general sufficient as an encoder. Firstly, the gradient of the loss inherently offers substantial information about data making it a good encoder. Secondly, a good latent space should satisfy local consistency \citep{zhou2003learning, kamnitsas2018semi}. GONs satisfy this since similar data points will have similar gradients due to the constant latent initialisation. As such, the network needs only to find an equilibrium where its prior is the gradient operation, allowing for significant freedom. Finally, since GONs are trained by backpropagating through empirical Bayes, there are benefits to using an activation function whose second derivative is non-zero.

%



\section{Results}
We evaluate Gradient Origin Networks on a variety of image datasets: MNIST \citep{lecun1998gradient}, Fashion-MNIST \citep{xiao2017fashion}, Small NORB \citep{lecun2004learning}, COIL-20 \citep{nane1996columbia}, CIFAR-10 \citep{krizhevsky2009learning}, CelebA \citep{liu2015faceattributes}, and LSUN Bedroom \citep{yu2015lsun}. Simple models are used: for small images, implicit GONs consist of approximately 4 hidden layers of 256 units and convolutional GONs consist of 4 convolution layers with Batch Normalization \citep{ioffe2015batch} and the ELU non-linearity \citep{clevert2015fast}, for larger images the same general architecture is used, scaled up with additional layers; all training is performed with the Adam optimiser \citep{Kingma2017AdamMethodStochastic}. Error bars in graphs represent standard deviations over three runs.


\begin{table}[H]
\centering
\begin{tabular}{lccccc}
                            & MNIST       & Fashion-MNIST & Small NORB  & COIL20      & CIFAR-10     \\ \hline
Single Step \\ \hline
GON (ours)                  & $\bf{0.41\!\pm\!0.01}$    & $\bf{1.00\!\pm\!0.01}$      & $\bf{0.17\!\pm\!0.01}$    & $5.35\!\pm\!0.01$        & $\bf{9.12\!\pm\!0.03}$         \\
AE                          & $1.33\!\pm\!0.02$        & $1.75\!\pm\!0.02$          & $0.73\!\pm\!0.01$        & $6.05\!\pm\!0.11$        & $12.24\!\pm\!0.05$        \\
Tied AE                  & $2.16\!\pm\!0.03$        & $2.45\!\pm\!0.02$          & $0.93\!\pm\!0.03$        & $6.68\!\pm\!0.13$        & $14.12\!\pm\!0.34$        \\
1 Step Detach             & $8.17\!\pm\!0.15$        & $7.76\!\pm\!0.21$          & $1.84\!\pm\!0.01$        & $15.72\!\pm\!0.70$       & $30.17\!\pm\!0.29$        \\
\hline 
Multiple Steps \\ \hline
10 Step Detach           & $8.13\!\pm\!0.50$        & $7.22\!\pm\!0.92$          & $1.78\!\pm\!0.02$        & $15.48\!\pm\!0.60$       & $28.68\!\pm\!1.16$        \\ 
10 Step                    & $0.42\!\pm\!0.01$        & $1.01\!\pm\!0.01$      & $\bf{0.17\!\pm\!0.01}$    & $5.36\!\pm\!0.01$        & $9.19\!\pm\!0.01$         \\ 
GLO                         & $0.70\!\pm\!0.01$        & $1.78\!\pm\!0.01$          & $0.61\!\pm\!0.01$        & $\bf{3.27\!\pm\!0.02}$    & $\bf{9.12\!\pm\!0.03}$  \\ 
\end{tabular}
\caption{Validation reconstruction loss (summed squared error) over 500 epochs. For GLO, latents are assigned to data points and jointly optimised with the network. GONs significantly outperform other single step methods and achieve the lowest reconstruction error on four of the five datasets.}
\label{tab:gon-recons}
\end{table}

\begin{table}[H]
\centering
\begin{tabular}{lcccccc}
             & MNIST     & Fashion-MNIST & Small NORB     & COIL20     & CIFAR-10 & CelebA   \\ \hline
VGON (ours)  & \bf 1.06  & 3.30          & \bf 2.34       & \bf 3.44   & \bf 5.85 & \bf 5.41 \\
VAE          & 1.15      & \bf 3.23      & 2.54           & 3.63       & 5.94     & 5.59     \\

\end{tabular}
\caption{Validation ELBO in bits/dim over 1000 epochs (CelebA is trained over 150 epochs).}
\label{tab:gon-vae}
\end{table}




\subsection{Quantitative Evaluation}

A quantitative evaluation of the representation ability of GONs is performed in Table~\ref{tab:gon-recons} against a number of baseline approaches. We compare against the single step methods: standard autoencoder, an autoencoder with tied weights \citep{gedeon1998stochastic}, and a GON with gradients detached from $\mathbf{z}$, as well as the multi-step methods: 10 gradient descent steps per data point, with and without detaching gradients, and a GLO \citep{bojanowski2018optimizing} which assigns a persistent latent vector to each data point and optimises them with gradient descent, therefore taking orders of magnitude longer to reconstruct validation data than other approaches. For the 10-step methods, a learning rate of 0.1 is applied as used in other literature \citep{nijkamp2019learning}; the GLO is trained with MSE for consistency with the other approaches and we do not project latents onto a hypersphere as proposed by the authors since in this experiment sampling is unimportant and this would handicap the approach. GONs achieve much lower validation loss than other single step methods and are competitive with the multi-step approaches; in fact, GONs achieve the lowest loss on four out of the five datasets.

\begin{figure}[b!]

\begin{subfigure}{0.32\textwidth}
\begin{adjustbox}{width=\linewidth}
\begin{tikzpicture}
\begin{axis}[
    width=7cm,height=6cm,
    xlabel={Training Epoch},
    ylabel={Validation Loss},
    xtick pos=left,
    ytick pos=left,
    enlarge x limits=false,
    every x tick/.style={color=black, thin},
    every y tick/.style={color=black, thin},
    tick align=outside,
    xlabel near ticks,
    ylabel near ticks,
    axis on top,
    legend style={draw=none},
    legend columns = 2,
]
\addplot+[spectral2, mark options={fill=spectral2}, mark repeat=5] table [x=Epoch, y=AETest, col sep=comma] {figures/gon-vs-ae/gon-vs-ae-1x1-nz=512,f=16,mean,eval,std.csv};\addlegendentry{AE 1x1}
\addplot+[spectral3, mark options={fill=spectral3}, mark repeat=5] table [x=Epoch, y=GONTest, col sep=comma] {figures/gon-vs-ae/gon-vs-ae-1x1-nz=512,f=16,mean,eval,std.csv};\addlegendentry{GON 1x1}

\addplot+[spectral10, mark options={fill=spectral10}, mark repeat=5] table [x=Epoch, y=AETest, col sep=comma] {figures/gon-vs-ae/gon-vs-ae-2x2-nz=512,f=16,mean,eval,std.csv};\addlegendentry{AE 2x2}
\addplot+[spectral4, mark options={fill=spectral4}, mark repeat=5] table [x=Epoch, y=GONTest, col sep=comma] {figures/gon-vs-ae/gon-vs-ae-2x2-nz=512,f=16,mean,eval,std.csv};\addlegendentry{GON 2x2}

\addplot+[spectral5, mark options={fill=spectral5}, mark repeat=5] table [x=Epoch, y=AETest, col sep=comma] {figures/gon-vs-ae/gon-vs-ae-4x4-nz=512,f=16,mean,eval,std.csv};\addlegendentry{AE 4x4}
\addplot+[spectral8, mark options={fill=spectral8}, mark repeat=5] table [x=Epoch, y=GONTest, col sep=comma] {figures/gon-vs-ae/gon-vs-ae-4x4-nz=512,f=16,mean,eval,std.csv};\addlegendentry{GON 4x4}

\addplot [name path=AE1upper,draw=none] table [x=Epoch,y expr=\thisrow{AETest}+\thisrow{AETestSTD}, col sep=comma] {figures/gon-vs-ae/gon-vs-ae-1x1-nz=512,f=16,mean,eval,std.csv};
\addplot [name path=AE1lower,draw=none] table [x=Epoch,y expr=\thisrow{AETest}-\thisrow{AETestSTD}, col sep=comma] {figures/gon-vs-ae/gon-vs-ae-1x1-nz=512,f=16,mean,eval,std.csv};
\addplot [fill=spectral2!20] fill between[of=AE1upper and AE1lower];

\addplot [name path=GON1upper,draw=none] table [x=Epoch,y expr=\thisrow{GONTest}+\thisrow{GONTestSTD}, col sep=comma] {figures/gon-vs-ae/gon-vs-ae-1x1-nz=512,f=16,mean,eval,std.csv};
\addplot [name path=GON1lower,draw=none] table [x=Epoch,y expr=\thisrow{GONTest}-\thisrow{GONTestSTD}, col sep=comma] {figures/gon-vs-ae/gon-vs-ae-1x1-nz=512,f=16,mean,eval,std.csv};
\addplot [fill=spectral3!20] fill between[of=GON1upper and GON1lower];

\addplot [name path=AE2upper,draw=none] table [x=Epoch,y expr=\thisrow{AETest}+\thisrow{AETestSTD}, col sep=comma] {figures/gon-vs-ae/gon-vs-ae-2x2-nz=512,f=16,mean,eval,std.csv};
\addplot [name path=AE2lower,draw=none] table [x=Epoch,y expr=\thisrow{AETest}-\thisrow{AETestSTD}, col sep=comma] {figures/gon-vs-ae/gon-vs-ae-2x2-nz=512,f=16,mean,eval,std.csv};
\addplot [fill=spectral10!20] fill between[of=AE2upper and AE2lower];

\addplot [name path=GON2upper,draw=none] table [x=Epoch,y expr=\thisrow{GONTest}+\thisrow{GONTestSTD}, col sep=comma] {figures/gon-vs-ae/gon-vs-ae-2x2-nz=512,f=16,mean,eval,std.csv};
\addplot [name path=GON2lower,draw=none] table [x=Epoch,y expr=\thisrow{GONTest}-\thisrow{GONTestSTD}, col sep=comma] {figures/gon-vs-ae/gon-vs-ae-2x2-nz=512,f=16,mean,eval,std.csv};
\addplot [fill=spectral4!20] fill between[of=GON2upper and GON2lower];

\addplot [name path=AE4upper,draw=none] table [x=Epoch,y expr=\thisrow{AETest}+\thisrow{AETestSTD}, col sep=comma] {figures/gon-vs-ae/gon-vs-ae-4x4-nz=512,f=16,mean,eval,std.csv};
\addplot [name path=AE4lower,draw=none] table [x=Epoch,y expr=\thisrow{AETest}-\thisrow{AETestSTD}, col sep=comma] {figures/gon-vs-ae/gon-vs-ae-4x4-nz=512,f=16,mean,eval,std.csv};
\addplot [fill=spectral5!20] fill between[of=AE4upper and AE4lower];

\addplot [name path=GON4upper,draw=none] table [x=Epoch,y expr=\thisrow{GONTest}+\thisrow{GONTestSTD}, col sep=comma] {figures/gon-vs-ae/gon-vs-ae-4x4-nz=512,f=16,mean,eval,std.csv};
\addplot [name path=GON4lower,draw=none] table [x=Epoch,y expr=\thisrow{GONTest}-\thisrow{GONTestSTD}, col sep=comma] {figures/gon-vs-ae/gon-vs-ae-4x4-nz=512,f=16,mean,eval,std.csv};
\addplot [fill=spectral8!20] fill between[of=GON4upper and GON4lower];

\end{axis}
\end{tikzpicture}
\end{adjustbox}
\caption{GONs achieve lower validation loss than autoencoders.}
\label{fig:gon-vs-ae-plot}
\end{subfigure}
\begin{subfigure}{0.32\textwidth}
\begin{adjustbox}{width=\linewidth}
\begin{tikzpicture}
\begin{axis}[
    width=7cm,height=6cm,
    xlabel={Training Epoch},
    ylabel={Validation Loss},
    xtick pos=left,
    ytick pos=left,
    enlarge x limits=false,
    every x tick/.style={color=black, thin},
    every y tick/.style={color=black, thin},
    tick align=outside,
    xlabel near ticks,
    ylabel near ticks,
    axis on top,
    legend style={draw=none},
    legend columns=2,
]
\addplot+[spectral2, mark options={fill=spectral2}, mark repeat=5] table [x=Epoch, y=1_step_test, col sep=comma] {figures/gon-multistep/multistep.csv};\addlegendentry{GON\textsuperscript{\textdagger}}

\addplot+[spectral3, mark options={fill=spectral3}, mark repeat=5] table [x=Epoch, y={1 step test}, col sep=comma] {figures/gon-multistep/detached_latents.csv};\addlegendentry{1 Step*}

\addplot+[spectral10, mark options={fill=spectral10}, mark repeat=5] table [x=Epoch, y=2_step_test, col sep=comma] {figures/gon-multistep/multistep.csv};\addlegendentry{2 Steps\textsuperscript{\textdagger}}

\addplot+[spectral5, mark options={fill=spectral5}, mark repeat=5] table [x=Epoch, y={10 step test}, col sep=comma] {figures/gon-multistep/detached_latents.csv};\addlegendentry{10 Steps*}


\addplot+[spectral8, mark options={fill=spectral8}, mark repeat=5] table [x=Epoch, y=8_step_test, col sep=comma] {figures/gon-multistep/multistep.csv};\addlegendentry{8 Steps\textsuperscript{\textdagger}}

\addplot+[spectral11, mark options={fill=spectral11}, mark repeat=5] table [x=Epoch, y={25 steps (lr=1e-2) test}, col sep=comma] {figures/gon-multistep/detached_latents.csv};\addlegendentry{25 Steps*}
\end{axis}
\end{tikzpicture}
\end{adjustbox}
\caption{Multiple latent update steps.\\ *=grads detached, \textsuperscript{\textdagger}=not detached.}
\label{fig:gon-multiple-steps-plot}
\end{subfigure}
\begin{subfigure}{0.32\textwidth}
\begin{adjustbox}{width=\linewidth}
\begin{tikzpicture}
\begin{axis}[
    width=7cm,height=6cm,
    xlabel={Training Epoch},
    ylabel={Reconstruction Loss},
    xtick pos=left,
    ytick pos=left,
    ymax=0.007,
    enlarge x limits=false,
    every x tick/.style={color=black, thin},
    every y tick/.style={color=black, thin},
    tick align=outside,
    xlabel near ticks,
    ylabel near ticks,
    axis on top,
    legend style={draw=none},
    legend columns=2,
]
\addplot+[spectral10, mark options={fill=spectral10}, mark repeat=50] table [x=Epoch, y={AETrain}, col sep=comma] {figures/long_plots/cifar_long.csv};\addlegendentry{AE Train}
\addplot+[spectral2, mark options={fill=spectral2}, mark repeat=50] table [x=Epoch, y={GON Train}, col sep=comma] {figures/long_plots/cifar_long.csv};\addlegendentry{GON Train}
\addplot+[spectral8, mark options={fill=spectral8}, mark repeat=50] table [x=Epoch, y={AE Test}, col sep=comma] {figures/long_plots/cifar_long.csv};\addlegendentry{AE Test}
\addplot+[spectral4, mark options={fill=spectral4}, mark repeat=50] table [x=Epoch, y={GON Test}, col sep=comma] {figures/long_plots/cifar_long.csv};\addlegendentry{GON Test}

\end{axis}
\end{tikzpicture}
\end{adjustbox}
\caption{GONs overfit less than standard autoencoders.}
\label{fig:gon-overfit-plot}
\end{subfigure}
\caption{Gradient Origin Networks trained on CIFAR-10 are found to outperform autoencoders using exactly the same architecture without the encoder, requiring half the number of parameters.}
\label{fig:gon-vs-ae}
\end{figure}
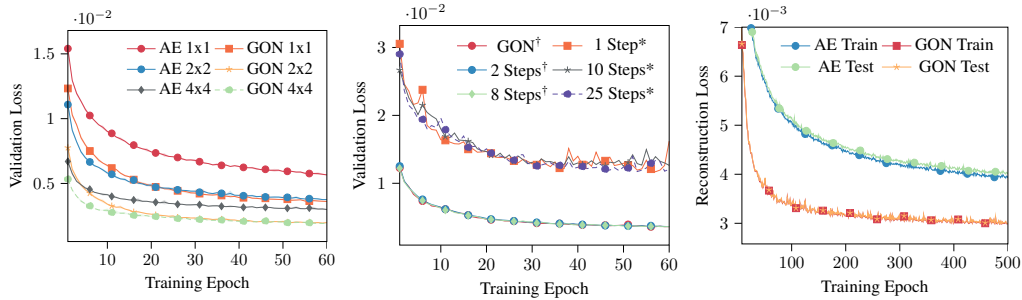

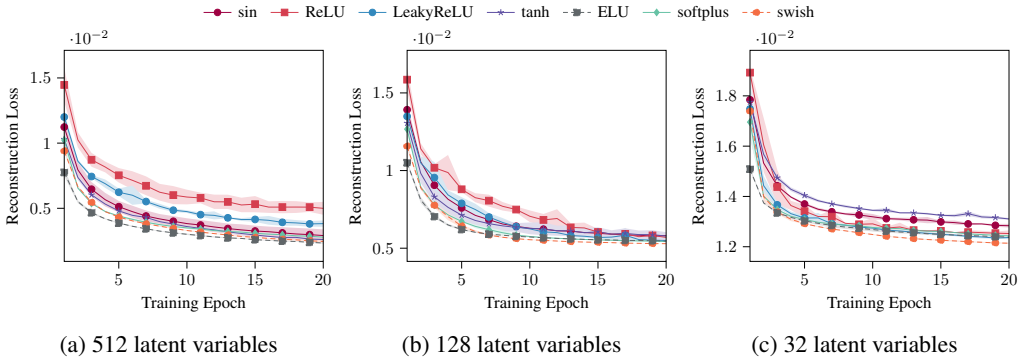
\begin{figure}[b!]
\centering

\begin{adjustbox}{width=0.6\textwidth}
\begin{tikzpicture}
\matrix[
        matrix of nodes,
        anchor=south,
        draw=none,
        inner sep=0.2em,
    ] {
        \ref{plots:sin}& sin&[4pt]
        \ref{plots:relu}& ReLU&[4pt]
        \ref{plots:leakyrelu}& LeakyReLU&[4pt]
        \ref{plots:tanh}& tanh&[4pt]
        \ref{plots:elu}& ELU&[4pt]
        \ref{plots:softplus}& softplus&[4pt]
        \ref{plots:swish}& swish&[4pt]\\
    };
\end{tikzpicture}
\end{adjustbox}

\begin{subfigure}{0.32\textwidth}
\centering
\begin{adjustbox}{width=\linewidth}
\begin{tikzpicture}
\begin{axis}[
    width=7cm,height=6cm,
    xlabel={Training Epoch},
    ylabel={Reconstruction Loss},
    xtick pos=left,
    ytick pos=left,
    enlarge x limits=false,
    every x tick/.style={color=black, thin},
    every y tick/.style={color=black, thin},
    tick align=outside,
    xlabel near ticks,
    ylabel near ticks,
    axis on top,
]
\addplot+[spectral1, mark options={fill=spectral1}, mark repeat=2] table [x=Epoch, y=sin, col sep=comma] {figures/activations/activations,cifar,nz=512,ngf=16,mean,std.csv};\addlegendentry{$\sin$}\label{plots:sin}
\addplot+[spectral2, mark options={fill=spectral2}, mark repeat=2] table [x=Epoch, y=RELU, col sep=comma] {figures/activations/activations,cifar,nz=512,ngf=16,mean,std.csv};\addlegendentry{ReLU}\label{plots:relu}
\addplot+[spectral10, mark options={fill=spectral10}, mark repeat=2] table [x=Epoch, y=leakyrelu, col sep=comma] {figures/activations/activations,cifar,nz=512,ngf=16,mean,std.csv};\addlegendentry{LeakyReLU}\label{plots:leakyrelu}
\addplot+[spectral11, mark options={fill=spectral11}, mark repeat=2] table [x=Epoch, y=tanh, col sep=comma] {figures/activations/activations,cifar,nz=512,ngf=16,mean,std.csv};\addlegendentry{tanh}\label{plots:tanh}
\addplot+[spectral9, mark options={fill=spectral9}, mark repeat=2] table [x=Epoch, y=softplus, col sep=comma] {figures/activations/activations,cifar,nz=512,ngf=16,mean,std.csv};\addlegendentry{softplus}\label{plots:softplus}
\addplot+[spectral3, mark options={fill=spectral3}, mark repeat=2] table [x=Epoch, y=swish, col sep=comma] {figures/activations/activations,cifar,nz=512,ngf=16,mean,std.csv};\addlegendentry{swish}\label{plots:swish}
\addplot+[spectral5, mark options={fill=spectral5}, mark repeat=2] table [x=Epoch, y=ELU, col sep=comma] {figures/activations/activations,cifar,nz=512,ngf=16,mean,std.csv};\addlegendentry{ELU}\label{plots:elu}
\legend{};

\addplot [name path=sinupper,draw=none] table [x=Epoch,y expr=\thisrow{sin}+\thisrow{sinstd}, col sep=comma] {figures/activations/activations,cifar,nz=512,ngf=16,mean,std.csv};
\addplot [name path=sinlower,draw=none] table [x=Epoch,y expr=\thisrow{sin}-\thisrow{sinstd}, col sep=comma] {figures/activations/activations,cifar,nz=512,ngf=16,mean,std.csv};
\addplot [fill=spectral1!20] fill between[of=sinupper and sinlower];

\addplot [name path=RELUupper,draw=none] table [x=Epoch,y expr=\thisrow{RELU}+\thisrow{RELUstd}, col sep=comma] {figures/activations/activations,cifar,nz=512,ngf=16,mean,std.csv};
\addplot [name path=RELUlower,draw=none] table [x=Epoch,y expr=\thisrow{RELU}-\thisrow{RELUstd}, col sep=comma] {figures/activations/activations,cifar,nz=512,ngf=16,mean,std.csv};
\addplot [fill=spectral2!20] fill between[of=RELUupper and RELUlower];

\addplot [name path=leakyreluupper,draw=none] table [x=Epoch,y expr=\thisrow{leakyrelu}+\thisrow{leakyrelustd}, col sep=comma] {figures/activations/activations,cifar,nz=512,ngf=16,mean,std.csv};
\addplot [name path=leakyrelulower,draw=none] table [x=Epoch,y expr=\thisrow{leakyrelu}-\thisrow{leakyrelustd}, col sep=comma] {figures/activations/activations,cifar,nz=512,ngf=16,mean,std.csv};
\addplot [fill=spectral10!20] fill between[of=leakyreluupper and leakyrelulower];

\addplot [name path=tanhupper,draw=none] table [x=Epoch,y expr=\thisrow{tanh}+\thisrow{tanhstd}, col sep=comma] {figures/activations/activations,cifar,nz=512,ngf=16,mean,std.csv};
\addplot [name path=tanhlower,draw=none] table [x=Epoch,y expr=\thisrow{tanh}-\thisrow{tanhstd}, col sep=comma] {figures/activations/activations,cifar,nz=512,ngf=16,mean,std.csv};
\addplot [fill=spectral11!20] fill between[of=tanhupper and tanhlower];

\addplot [name path=softplusupper,draw=none] table [x=Epoch,y expr=\thisrow{softplus}+\thisrow{softplusstd}, col sep=comma] {figures/activations/activations,cifar,nz=512,ngf=16,mean,std.csv};
\addplot [name path=softpluslower,draw=none] table [x=Epoch,y expr=\thisrow{softplus}-\thisrow{softplusstd}, col sep=comma] {figures/activations/activations,cifar,nz=512,ngf=16,mean,std.csv};
\addplot [fill=spectral9!20] fill between[of=softplusupper and softpluslower];

\addplot [name path=swishupper,draw=none] table [x=Epoch,y expr=\thisrow{swish}+\thisrow{swishstd}, col sep=comma] {figures/activations/activations,cifar,nz=512,ngf=16,mean,std.csv};
\addplot [name path=swishlower,draw=none] table [x=Epoch,y expr=\thisrow{swish}-\thisrow{swishstd}, col sep=comma] {figures/activations/activations,cifar,nz=512,ngf=16,mean,std.csv};
\addplot [fill=spectral3!20] fill between[of=swishupper and swishlower];

\addplot [name path=ELUupper,draw=none] table [x=Epoch,y expr=\thisrow{ELU}+\thisrow{ELUstd}, col sep=comma] {figures/activations/activations,cifar,nz=512,ngf=16,mean,std.csv};
\addplot [name path=ELUlower,draw=none] table [x=Epoch,y expr=\thisrow{ELU}-\thisrow{ELUstd}, col sep=comma] {figures/activations/activations,cifar,nz=512,ngf=16,mean,std.csv};
\addplot [fill=spectral5!20] fill between[of=ELUupper and ELUlower];

\end{axis}
\coordinate (top) at (rel axis cs:0,1);
\coordinate (bot) at (rel axis cs:1,0);
\path (top|-current bounding box.north)--
      coordinate(legendpos)
      (bot|-current bounding box.north);
\end{tikzpicture}
\end{adjustbox}
\caption{512 latent variables}
\end{subfigure}
\begin{subfigure}{0.32\textwidth}
\centering
\begin{adjustbox}{width=\linewidth}
\begin{tikzpicture}
\begin{axis}[
    width=7cm,height=6cm,
    xlabel={Training Epoch},
    ylabel={Reconstruction Loss},
    xtick pos=left,
    ytick pos=left,
    enlarge x limits=false,
    every x tick/.style={color=black, thin},
    every y tick/.style={color=black, thin},
    tick align=outside,
    xlabel near ticks,
    ylabel near ticks,
    axis on top,
]
\addplot+[spectral1, mark options={fill=spectral1}, mark repeat=2] table [x=Epoch, y=sin, col sep=comma] {figures/activations/activations,cifar,nz=128,ngf=16,mean,std.csv};\addlegendentry{$\sin$}
\addplot+[spectral2, mark options={fill=spectral2}, mark repeat=2] table [x=Epoch, y=RELU, col sep=comma] {figures/activations/activations,cifar,nz=128,ngf=16,mean,std.csv};\addlegendentry{ReLU}
\addplot+[spectral10, mark options={fill=spectral10}, mark repeat=2] table [x=Epoch, y=leakyrelu, col sep=comma] {figures/activations/activations,cifar,nz=128,ngf=16,mean,std.csv};\addlegendentry{LeakyReLU}
\addplot+[spectral11, mark options={fill=spectral11}, mark repeat=2] table [x=Epoch, y=tanh, col sep=comma] {figures/activations/activations,cifar,nz=128,ngf=16,mean,std.csv};\addlegendentry{tanh}
\addplot+[spectral9, mark options={fill=spectral9}, mark repeat=2] table [x=Epoch, y=softplus, col sep=comma] {figures/activations/activations,cifar,nz=128,ngf=16,mean,std.csv};\addlegendentry{softplus}
\addplot+[spectral3, mark options={fill=spectral3}, mark repeat=2] table [x=Epoch, y=swish, col sep=comma] {figures/activations/activations,cifar,nz=128,ngf=16,mean,std.csv};\addlegendentry{swish}
\addplot+[spectral5, mark options={fill=spectral5}, mark repeat=2] table [x=Epoch, y=ELU, col sep=comma] {figures/activations/activations,cifar,nz=128,ngf=16,mean,std.csv};\addlegendentry{ELU}
\legend{};

\addplot [name path=sinupper,draw=none] table [x=Epoch,y expr=\thisrow{sin}+\thisrow{sinstd}, col sep=comma] {figures/activations/activations,cifar,nz=128,ngf=16,mean,std.csv};
\addplot [name path=sinlower,draw=none] table [x=Epoch,y expr=\thisrow{sin}-\thisrow{sinstd}, col sep=comma] {figures/activations/activations,cifar,nz=128,ngf=16,mean,std.csv};
\addplot [fill=spectral1!20] fill between[of=sinupper and sinlower];

\addplot [name path=RELUupper,draw=none] table [x=Epoch,y expr=\thisrow{RELU}+\thisrow{RELUstd}, col sep=comma] {figures/activations/activations,cifar,nz=128,ngf=16,mean,std.csv};
\addplot [name path=RELUlower,draw=none] table [x=Epoch,y expr=\thisrow{RELU}-\thisrow{RELUstd}, col sep=comma] {figures/activations/activations,cifar,nz=128,ngf=16,mean,std.csv};
\addplot [fill=spectral2!20] fill between[of=RELUupper and RELUlower];

\addplot [name path=leakyreluupper,draw=none] table [x=Epoch,y expr=\thisrow{leakyrelu}+\thisrow{leakyrelustd}, col sep=comma] {figures/activations/activations,cifar,nz=128,ngf=16,mean,std.csv};
\addplot [name path=leakyrelulower,draw=none] table [x=Epoch,y expr=\thisrow{leakyrelu}-\thisrow{leakyrelustd}, col sep=comma] {figures/activations/activations,cifar,nz=128,ngf=16,mean,std.csv};
\addplot [fill=spectral10!20] fill between[of=leakyreluupper and leakyrelulower];

\addplot [name path=tanhupper,draw=none] table [x=Epoch,y expr=\thisrow{tanh}+\thisrow{tanhstd}, col sep=comma] {figures/activations/activations,cifar,nz=128,ngf=16,mean,std.csv};
\addplot [name path=tanhlower,draw=none] table [x=Epoch,y expr=\thisrow{tanh}-\thisrow{tanhstd}, col sep=comma] {figures/activations/activations,cifar,nz=128,ngf=16,mean,std.csv};
\addplot [fill=spectral11!20] fill between[of=tanhupper and tanhlower];

\addplot [name path=softplusupper,draw=none] table [x=Epoch,y expr=\thisrow{softplus}+\thisrow{softplusstd}, col sep=comma] {figures/activations/activations,cifar,nz=128,ngf=16,mean,std.csv};
\addplot [name path=softpluslower,draw=none] table [x=Epoch,y expr=\thisrow{softplus}-\thisrow{softplusstd}, col sep=comma] {figures/activations/activations,cifar,nz=128,ngf=16,mean,std.csv};
\addplot [fill=spectral9!20] fill between[of=softplusupper and softpluslower];

\addplot [name path=swishupper,draw=none] table [x=Epoch,y expr=\thisrow{swish}+\thisrow{swishstd}, col sep=comma] {figures/activations/activations,cifar,nz=128,ngf=16,mean,std.csv};
\addplot [name path=swishlower,draw=none] table [x=Epoch,y expr=\thisrow{swish}-\thisrow{swishstd}, col sep=comma] {figures/activations/activations,cifar,nz=128,ngf=16,mean,std.csv};
\addplot [fill=spectral3!20] fill between[of=swishupper and swishlower];

\addplot [name path=ELUupper,draw=none] table [x=Epoch,y expr=\thisrow{ELU}+\thisrow{ELUstd}, col sep=comma] {figures/activations/activations,cifar,nz=128,ngf=16,mean,std.csv};
\addplot [name path=ELUlower,draw=none] table [x=Epoch,y expr=\thisrow{ELU}-\thisrow{ELUstd}, col sep=comma] {figures/activations/activations,cifar,nz=128,ngf=16,mean,std.csv};
\addplot [fill=spectral5!20] fill between[of=ELUupper and ELUlower];

\end{axis}
\end{tikzpicture}
\end{adjustbox}
\caption{128 latent variables}
\end{subfigure}
\begin{subfigure}{0.32\textwidth}
\centering
\begin{adjustbox}{width=\linewidth}
\begin{tikzpicture}
\begin{axis}[
    width=7cm,height=6cm,
    xlabel={Training Epoch},
    ylabel={Reconstruction Loss},
    xtick pos=left,
    ytick pos=left,
    enlarge x limits=false,
    every x tick/.style={color=black, thin},
    every y tick/.style={color=black, thin},
    tick align=outside,
    xlabel near ticks,
    ylabel near ticks,
    axis on top,
]
\addplot+[spectral1, mark options={fill=spectral1}, mark repeat=2] table [x=Epoch, y=sin, col sep=comma] {figures/activations/activations,cifar,nz=32,ngf=16,mean,std.csv};\addlegendentry{$\sin$}
\addplot+[spectral2, mark options={fill=spectral2}, mark repeat=2] table [x=Epoch, y=RELU, col sep=comma] {figures/activations/activations,cifar,nz=32,ngf=16,mean,std.csv};\addlegendentry{ReLU}
\addplot+[spectral10, mark options={fill=spectral10}, mark repeat=2] table [x=Epoch, y=leakyrelu, col sep=comma] {figures/activations/activations,cifar,nz=32,ngf=16,mean,std.csv};\addlegendentry{LeakyReLU}
\addplot+[spectral11, mark options={fill=spectral11}, mark repeat=2] table [x=Epoch, y=tanh, col sep=comma] {figures/activations/activations,cifar,nz=32,ngf=16,mean,std.csv};\addlegendentry{tanh}
\addplot+[spectral9, mark options={fill=spectral9}, mark repeat=2] table [x=Epoch, y=softplus, col sep=comma] {figures/activations/activations,cifar,nz=32,ngf=16,mean,std.csv};\addlegendentry{softplus}
\addplot+[spectral3, mark options={fill=spectral3}, mark repeat=2] table [x=Epoch, y=swish, col sep=comma] {figures/activations/activations,cifar,nz=32,ngf=16,mean,std.csv};\addlegendentry{swish}
\addplot+[spectral5, mark options={fill=spectral5}, mark repeat=2] table [x=Epoch, y=ELU, col sep=comma] {figures/activations/activations,cifar,nz=32,ngf=16,mean,std.csv};\addlegendentry{ELU}
\legend{};

\addplot [name path=sinupper,draw=none] table [x=Epoch,y expr=\thisrow{sin}+\thisrow{sinstd}, col sep=comma] {figures/activations/activations,cifar,nz=32,ngf=16,mean,std.csv};
\addplot [name path=sinlower,draw=none] table [x=Epoch,y expr=\thisrow{sin}-\thisrow{sinstd}, col sep=comma] {figures/activations/activations,cifar,nz=32,ngf=16,mean,std.csv};
\addplot [fill=spectral1!20] fill between[of=sinupper and sinlower];

\addplot [name path=RELUupper,draw=none] table [x=Epoch,y expr=\thisrow{RELU}+\thisrow{RELUstd}, col sep=comma] {figures/activations/activations,cifar,nz=32,ngf=16,mean,std.csv};
\addplot [name path=RELUlower,draw=none] table [x=Epoch,y expr=\thisrow{RELU}-\thisrow{RELUstd}, col sep=comma] {figures/activations/activations,cifar,nz=32,ngf=16,mean,std.csv};
\addplot [fill=spectral2!20] fill between[of=RELUupper and RELUlower];

\addplot [name path=leakyreluupper,draw=none] table [x=Epoch,y expr=\thisrow{leakyrelu}+\thisrow{leakyrelustd}, col sep=comma] {figures/activations/activations,cifar,nz=32,ngf=16,mean,std.csv};
\addplot [name path=leakyrelulower,draw=none] table [x=Epoch,y expr=\thisrow{leakyrelu}-\thisrow{leakyrelustd}, col sep=comma] {figures/activations/activations,cifar,nz=32,ngf=16,mean,std.csv};
\addplot [fill=spectral10!20] fill between[of=leakyreluupper and leakyrelulower];

\addplot [name path=tanhupper,draw=none] table [x=Epoch,y expr=\thisrow{tanh}+\thisrow{tanhstd}, col sep=comma] {figures/activations/activations,cifar,nz=32,ngf=16,mean,std.csv};
\addplot [name path=tanhlower,draw=none] table [x=Epoch,y expr=\thisrow{tanh}-\thisrow{tanhstd}, col sep=comma] {figures/activations/activations,cifar,nz=32,ngf=16,mean,std.csv};
\addplot [fill=spectral11!20] fill between[of=tanhupper and tanhlower];

\addplot [name path=softplusupper,draw=none] table [x=Epoch,y expr=\thisrow{softplus}+\thisrow{softplusstd}, col sep=comma] {figures/activations/activations,cifar,nz=32,ngf=16,mean,std.csv};
\addplot [name path=softpluslower,draw=none] table [x=Epoch,y expr=\thisrow{softplus}-\thisrow{softplusstd}, col sep=comma] {figures/activations/activations,cifar,nz=32,ngf=16,mean,std.csv};
\addplot [fill=spectral9!20] fill between[of=softplusupper and softpluslower];

\addplot [name path=swishupper,draw=none] table [x=Epoch,y expr=\thisrow{swish}+\thisrow{swishstd}, col sep=comma] {figures/activations/activations,cifar,nz=32,ngf=16,mean,std.csv};
\addplot [name path=swishlower,draw=none] table [x=Epoch,y expr=\thisrow{swish}-\thisrow{swishstd}, col sep=comma] {figures/activations/activations,cifar,nz=32,ngf=16,mean,std.csv};
\addplot [fill=spectral3!20] fill between[of=swishupper and swishlower];

\addplot [name path=ELUupper,draw=none] table [x=Epoch,y expr=\thisrow{ELU}+\thisrow{ELUstd}, col sep=comma] {figures/activations/activations,cifar,nz=32,ngf=16,mean,std.csv};
\addplot [name path=ELUlower,draw=none] table [x=Epoch,y expr=\thisrow{ELU}-\thisrow{ELUstd}, col sep=comma] {figures/activations/activations,cifar,nz=32,ngf=16,mean,std.csv};
\addplot [fill=spectral5!20] fill between[of=ELUupper and ELUlower];
\end{axis}
\end{tikzpicture}
\end{adjustbox}
\caption{32 latent variables}
\end{subfigure}

\caption{The impact of activation function and number of latent variables on model performance for a GON measured by comparing reconstruction losses through training.}
\label{fig:gon-activations}
\end{figure}

Our variational GON is compared with a VAE, whose decoder is the same as the GON, quantitatively in terms of ELBO on the test set in Table \ref{tab:gon-vae}. We find that the representation ability of GONs is pertinent here also, allowing the variational GON to achieve lower ELBO on five out of the six datasets tested. Both models were trained with the original VAE formulation for fairness, however, we note that variations aimed at improving VAE sample quality such as $\beta$-VAEs \citep{higgins2017beta} are also applicable to variational GONs to the same effect.

An ablation study is performed, comparing convolutional GONs with autoencoders whose decoders have exactly the same architecture as the GONs (Figure~\ref{fig:gon-vs-ae-plot}) and where the autoencoder decoders mirror their respective encoders. The mean reconstruction loss over the test set is measured after each epoch for a variety of latent sizes. Despite having half the number of parameters and linear encodings, GONs achieve significantly lower reconstruction loss over a wide range of architectures. Further experiments evaluating the impact of latent size and model capacity are performed in Appendix~\ref{apx:compare-autoencoders}.

Our hypothesis that for high dimensional datasets, a single gradient step is sufficient when jointly optimised with the forwards pass is tested on the CIFAR-10 dataset in Figure~\ref{fig:gon-multiple-steps-plot}. We observe negligible difference between a single step and multiple jointly optimised steps, in support of our hypothesis. Performing multiple steps greatly increases run-time so there is seemingly no benefit in this case. Additionally, the importance of the joint optimisation procedure is determined by detaching the gradients from $\mathbf{z}$ before reconstructing (Figure~\ref{fig:gon-multiple-steps-plot}); this results in markedly worse performance, even when in conjunction with multiple steps. In Figure~\ref{fig:gon-overfit-plot} we assess whether the greater performance of GONs relative to autoencoders comes at the expense of generalisation; we find that the opposite is true, that the discrepancy between reconstruction loss on the training and test sets is greater with autoencoders. This result extends to other datasets as can be seen in Appendix~\ref{apx:gon-generalisation}.

Figure~\ref{fig:gon-activations} demonstrates the effect activation functions have on convolutional GON performance for different numbers of latent variables. Since optimising GONs requires computing second order derivatives, the choice of nonlinearity requires different characteristics to standard models. In particular, GONs prefer functions that are not only effective activation functions, but also whose second derivatives are non-zero, unlike ReLUs where $\text{ReLU}''(x)=0$. The ELU non-linearity is effective with all tested architectures.




\begin{figure}[b!]
    \centering
    \begin{subfigure}[t]{0.19\textwidth}
        \centering
        \includegraphics[width=\linewidth]{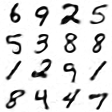}
        \caption{4,385 parameters}
        \label{fig:tiny-overfit}
    \end{subfigure}
    \hfill
    \begin{subfigure}[t]{0.19\textwidth}
        \centering
        \includegraphics[width=\linewidth]{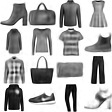}
        \caption{297k parameters}
    \end{subfigure}
    \hfill
    \begin{subfigure}[t]{0.19\textwidth}
       \centering
       \includegraphics[width=\linewidth]{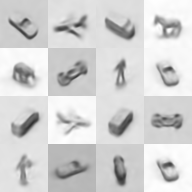}
        \caption{270k parameters}
    \end{subfigure}
    \hfill
    \begin{subfigure}[t]{0.19\textwidth}
       \centering
       \includegraphics[width=\linewidth]{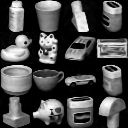}
        \caption{270k parameters}
    \end{subfigure}
    \hfill
    \begin{subfigure}[t]{0.19\textwidth}
       \centering
       \includegraphics[width=\linewidth]{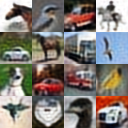}
        \caption{396k parameters}
    \end{subfigure}
\caption{Training implicit GONs with few parameters demonstrates their representation ability.}
\label{fig:fitting}
\end{figure}

\begin{figure}[b!]
    \centering
    
    \begin{tikzpicture}
    \node (fig) {\includegraphics[width=0.99\linewidth]{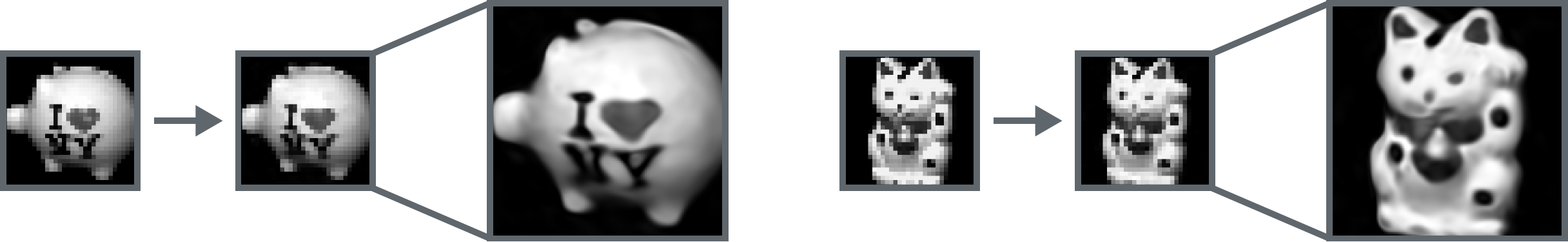}};
    \node (otxt) at (-6.32,0.9) {original};
    \node (gtxt) at (-4.26,0.9) {GON};
    \node (otxt2) at (1.1,0.9) {original};
    \node (gtxt2) at (3.15,0.9) {GON};
    \end{tikzpicture}

    \caption{By training an implicit GON on 32x32 images, then sampling at 256x256, super-resolution is possible despite never observing high resolution data.}
    \label{fig:supersample-coil-cherry}
\end{figure}
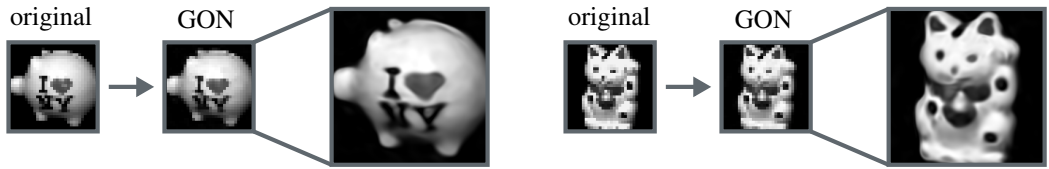

\subsection{Qualitative Evaluation}

\begin{figure}[b!]
    \centering
    \begin{subfigure}[t]{0.19\textwidth}
        \centering
        \includegraphics[width=\linewidth]{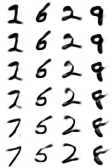}
        \caption{MNIST}
    \end{subfigure}
    \hfill
    \begin{subfigure}[t]{0.19\textwidth}
        \centering
        \includegraphics[width=\linewidth]{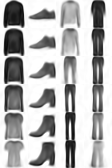}
        \caption{Fashion-MNIST}
        \label{fig:fashion-slerp}
    \end{subfigure}
    \hfill
    \begin{subfigure}[t]{0.19\textwidth}
       \centering
       \includegraphics[width=\linewidth]{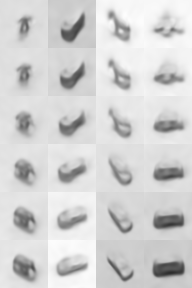}
        \caption{Small NORB}
    \end{subfigure}
    \hfill
    \begin{subfigure}[t]{0.19\textwidth}
       \centering
       \includegraphics[width=\linewidth]{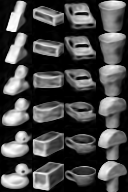}
        \caption{COIL20}
        \label{fig:coil-slerp}
    \end{subfigure}
    \hfill
    \begin{subfigure}[t]{0.19\textwidth}
       \centering
       \includegraphics[width=\linewidth]{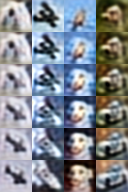}
        \caption{CIFAR-10}
    \end{subfigure}
\caption{Spherical linear interpolations between points in the latent space for trained implicit GONs using different datasets (approximately 2-10 minutes training per dataset on a single GPU).}
\label{fig:interpolations}
\end{figure}

The representation ability of implicit GONs is shown in Figure~\ref{fig:fitting} where we train on large image datasets using a relatively small number of parameters. In particular, Figure~\ref{fig:tiny-overfit} shows MNIST can be well-represented with just 4,385 parameters (a SIREN with 3 hidden layers each with 32 hidden units, and 32 latent dimensions).
An advantage of modelling data with implicit networks is that coordinates can be at arbitrarily high resolutions. In Figure~\ref{fig:supersample-coil-cherry} we train on 32x32 images then reconstruct at 256x256. A significant amount of high frequency detail is modelled despite only seeing low resolution images. 
The structure of the implicit GON latent space is shown by sampling latent codes from pairs of images, and then spherically interpolating between them to synthesise new samples (Figure~\ref{fig:interpolations}). These samples are shown to capture variations in shape (the shoes in Figure~\ref{fig:fashion-slerp}), size, and rotation (Figure~\ref{fig:coil-slerp}).

\begin{figure}[b!]
    \centering
    \begin{subfigure}[t]{0.19\textwidth}
        \centering
        \includegraphics[width=\linewidth]{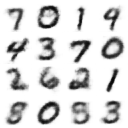}
        \caption{4,331 parameters}
    \end{subfigure}
    \hfill
    \begin{subfigure}[t]{0.19\textwidth}
       \centering
       \includegraphics[width=\linewidth]{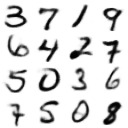}
        \caption{270k parameters}
    \end{subfigure}
    \hfill
    \begin{subfigure}[t]{0.19\textwidth}
        \centering
        \includegraphics[width=\linewidth]{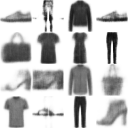}
        \caption{270k parameters}
    \end{subfigure}
    \hfill
    \begin{subfigure}[t]{0.19\textwidth}
       \centering
       \includegraphics[width=\linewidth]{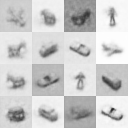}
        \caption{270k parameters}
    \end{subfigure}
    \hfill
    \begin{subfigure}[t]{0.19\textwidth}
       \centering
       \includegraphics[width=\linewidth]{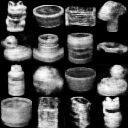}
        \caption{270k parameters}
    \end{subfigure}
\caption{Random samples from a convolutional variational GON with normally distributed latents.}
\label{fig:conv-samples}
\end{figure}

\begin{figure}[b!]
    \centering
    \begin{subfigure}[t]{0.19\textwidth}
        \centering
        \includegraphics[width=\linewidth]{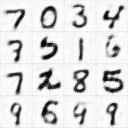}
        \caption{3s 300 steps}
    \end{subfigure}
    \hfill
    \begin{subfigure}[t]{0.19\textwidth}
        \centering
        \includegraphics[width=\linewidth]{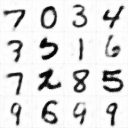}
        \caption{10s 1000 steps}
    \end{subfigure}
    \hfill
    \begin{subfigure}[t]{0.19\textwidth}
       \centering
       \includegraphics[width=\linewidth]{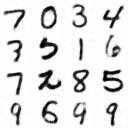}
        \caption{30s 3000 steps}
    \end{subfigure}
    \hfill
    \begin{subfigure}[t]{0.19\textwidth}
       \centering
       \includegraphics[width=\linewidth]{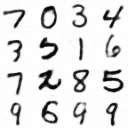}
        \caption{60s 6000 steps}
    \end{subfigure}
    \hfill
    \begin{subfigure}[t]{0.19\textwidth}
       \centering
       \includegraphics[width=\linewidth]{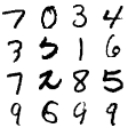}
        \caption{Ground Truth}
    \end{subfigure}
\caption{Convergence of convolutional GONs with 74k parameters.}
\label{fig:mnist-training}
\end{figure}

We assess the quality of samples from variational GONs using convolutional models trained to convergence in Figure~\ref{fig:conv-samples}. These are diverse and often contain fine details. Samples from an implicit GON trained with early stopping, as a simple alternative, can be found in Appendix~\ref{apx:early-stopping} however this approach results in fine details being lost.
GONs are also found to converge quickly; we plot reconstructions at multiple time points during the first minute of training (Figure~\ref{fig:mnist-training}). After only 3 seconds of training on a single GPU, a large amount of signal information from MNIST is modelled.

\begin{figure}[h]
    \centering
    \begin{subfigure}[t]{0.65306122449\textwidth}
        \centering
        \includegraphics[width=0.49\linewidth,trim={0 0 0 0},clip]{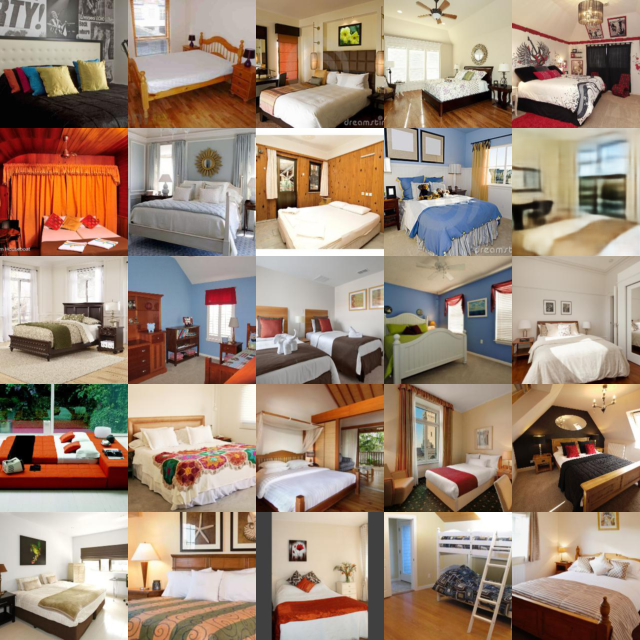}
        \hfill
        \includegraphics[width=0.49\linewidth,trim={0 0 0 0},clip]{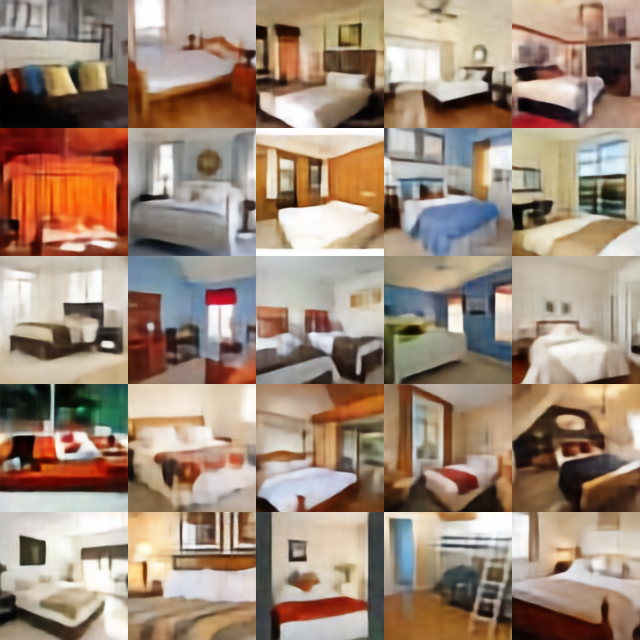}
        \caption{LSUN 128x128 Bedroom validation images (left) reconstructed by a convolutional GON (right).}
        \label{fig:lsun-representation}
    \end{subfigure}
    \hfill
    \begin{subfigure}[t]{0.32\textwidth}
        \centering
        \includegraphics[width=\linewidth,trim={0 0 0 0},clip]{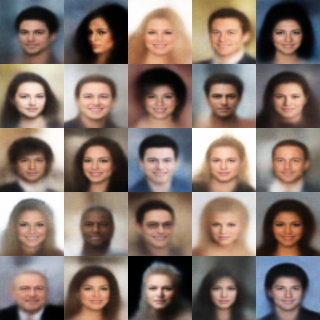}
        \caption{Samples from a convolutional variational GON trained on CelebA.}
        \label{fig:celeba-vgon}
    \end{subfigure}
    \caption{GONs are able to represent high resolution complex datasets to a high degree of fidelity.}
\end{figure}

In order to evaluate how well GONs can represent high resolution natural images, we train a convolutional GON on the LSUN Bedroom dataset scaled to 128x128 (Figure~\ref{fig:lsun-representation}). As with smaller, more simple data, we find training to always be extremely stable and consistent over a wide range of hyperparameter settings. Reconstructions are of excellent quality given the simple network architecture.
A convolutional variational GON is also trained on the CelebA dataset scaled to 64x64 (Figure~\ref{fig:celeba-vgon}). Unconditional samples are somewhat blurry as commonly associated with traditional VAE models on complex natural images \citep{zhao2017towards} but otherwise show wide variety.

\section{Discussion}

Despite similarities with autoencoder approaches, the absence of an encoding network offers several advantages. VAEs with overpowered decoders are known to ignore the latent variables \citep{Chen2017VariationalLossyAutoencoder} whereas GONs only have one network that equally serves both encoding and decoding functionality. Designing inference networks for complicated decoders is not a trivial task \citep{vahdat2020NVAE}, however, inferring latent variables using a GON simplifies this procedure. Similar to GONs, Normalizing Flow methods are also capable of encoding and decoding with a single set of weights, however, they achieve this by restricting the network to be invertible. This requires considerable architectural restrictions that affect performance, make them less parameter efficient, and are unable to reduce the dimension of the data \citep{kingma2018glow}. Similarly, autoencoders with tied weights also encode and decode with a single set of weights by using the transpose of the encoder's weight matrices in the decoder; this, however, is only applicable to simple architectures. GONs on the other hand use gradients as encodings which allow arbitrary functions to be used.

A number of previous works have used gradient-based computations to learn latent vectors however as far as we are aware, we are the first to use a single gradient step jointly optimised with the feedforward pass, making it fundamentally different to these approaches. Latent encodings have been estimated for pre-trained generative models without encoders, namely Generative Adversarial Networks, using approaches such as standard gradient descent \citep{lipton2017precise, zhu2016generative}. A number of approaches have trained generative models directly with gradient-based inference \citep{Han2017Alternating, bojanowski2018optimizing, Zadeh2019VariationalAutoDecoder}; these assign latent vectors to data points and jointly learn them with the network parameters through standard gradient descent or Langevin dynamics. This is very slow, however, and convergence for unseen data samples is not guaranteed. Short run MCMC has also been applied \citep{Nijkamp2020LearningMultilayerLatent} however this still requires approximately 25 update steps. Since GONs train end-to-end, the optimiser can make use of the second order derivatives to allow for inference in a single step. Also of relevance is model-agnostic meta-learning \citep{finn2017model}, which trains an architecture so that a few gradient descent steps are all that are necessary to new tasks. This is achieved by backpropagating through these gradients, similar to GONs.



In the case of implicit GONs, the integration terms in Equation \ref{eqn:implicit-gon-loss} result in computation time that scales in proportion with the data dimension. This makes training slower for very high dimensional data, although we have not yet investigated Monte Carlo integration. In general GONs are stable and consistent, capable of generating quality samples with an exceptionally small number of parameters, and converge to diverse results with few iterations. Nevertheless, there are avenues to explore so as to improve the quality of samples and scale to larger datasets. In particular, it would be beneficial to focus on how to better sample these models, perform formal analysis on the gradients, and investigate whether the distance function could be improved to better capture fine details.


\section*{Conclusion}


In conclusion, we have introduced a method based on empirical Bayes which computes the gradient of the data fitting loss with respect to the origin, and then jointly fits the data while learning this new point of reference in the latent space. The results show that this approach is able to represent datasets using a small number of parameters with a simple overall architecture, which has advantages in applications such as implicit representation networks. GONs are shown to converge faster with lower overall reconstruction loss than autoencoders, using the exact same architecture but without the encoder. Experiments show that the choice of non-linearity is important, as the network derivative jointly acts as the encoding function.

\bibliography{references_new}

\begin{thebibliography}{47}
\providecommand{\natexlab}[1]{#1}
\providecommand{\url}[1]{\texttt{#1}}
\expandafter\ifx\csname urlstyle\endcsname\relax
  \providecommand{\doi}[1]{doi: #1}\else
  \providecommand{\doi}{doi: \begingroup \urlstyle{rm}\Url}\fi

\bibitem[Bojanowski et~al.(2018)Bojanowski, Joulin, Lopez-Pas, and
  Szlam]{bojanowski2018optimizing}
Piotr Bojanowski, Armand Joulin, David Lopez-Pas, and Arthur Szlam.
\newblock Optimizing the {Latent} {Space} of {Generative} {Networks}.
\newblock In \emph{Proceedings of the 35th International Conference on Machine
  Learning, {ICML}}, volume~80, pp.\  600--609, 2018.

\bibitem[Chen et~al.(2020)Chen, Radford, Child, Wu, Jun, Dhariwal, Luan, and
  Sutskever]{chen2020generative}
Mark Chen, Alec Radford, Rewon Child, Jeff Wu, Heewoo Jun, Prafulla Dhariwal,
  David Luan, and Ilya Sutskever.
\newblock Generative {Pretraining} from {Pixels}.
\newblock In \emph{Proceedings of the 37th International Conference on Machine
  Learning}, 2020.

\bibitem[Chen et~al.(2017)Chen, Kingma, Salimans, Duan, Dhariwal, Schulman,
  Sutskever, and Abbeel]{Chen2017VariationalLossyAutoencoder}
Xi~Chen, Diederik~P. Kingma, Tim Salimans, Yan Duan, Prafulla Dhariwal, John
  Schulman, Ilya Sutskever, and Pieter Abbeel.
\newblock Variational {Lossy} {Autoencoder}.
\newblock In \emph{5th International {Conference} on {Learning}
  {Representations}, ICLR}, 2017.

\bibitem[Clevert et~al.(2016)Clevert, Unterthiner, and
  Hochreiter]{clevert2015fast}
Djork{-}Arn{\'{e}} Clevert, Thomas Unterthiner, and Sepp Hochreiter.
\newblock Fast and {Accurate} {Deep} {Network} {Learning} by {Exponential}
  {Linear} {Units} ({ELUs}).
\newblock In \emph{4th International Conference on Learning Representations,
  {ICLR}}, 2016.

\bibitem[Dai et~al.(2019)Dai, Cai, Zhang, Xia, and Zhang]{dai2019second}
Tao Dai, Jianrui Cai, Yongbing Zhang, Shu-Tao Xia, and Lei Zhang.
\newblock Second-order {Attention} {Network} for {Single} {Image}
  {Super}-{Resolution}.
\newblock In \emph{Proceedings of the IEEE Conference on Computer Vision and
  Pattern Recognition}, pp.\  11065--11074, 2019.

\bibitem[Du \& Mordatch(2019)Du and
  Mordatch]{Du2019ImplicitGenerationGeneralization}
Yilun Du and Igor Mordatch.
\newblock Implicit {Generation} and {Generalization} in {Energy}-{Based}
  {Models}.
\newblock In \emph{Neural Information Processing Systems (NeurIPS)}, 2019.

\bibitem[Fefferman et~al.(2016)Fefferman, Mitter, and
  Narayanan]{fefferman2016testing}
Charles Fefferman, Sanjoy Mitter, and Hariharan Narayanan.
\newblock Testing the {Manifold} {Hypothesis}.
\newblock \emph{Journal of the American Mathematical Society}, 29\penalty0
  (4):\penalty0 983--1049, 2016.

\bibitem[Finn et~al.(2017)Finn, Abbeel, and Levine]{finn2017model}
Chelsea Finn, Pieter Abbeel, and Sergey Levine.
\newblock Model-{Agnostic} {Meta}-{Learning} for {Fast} {Adaptation} of {Deep}
  {Networks}.
\newblock \emph{Proceedings of the 34th International Conference on Machine
  Learning}, 2017.

\bibitem[Gedeon(1998)]{gedeon1998stochastic}
TD~Gedeon.
\newblock Stochastic {Bidirectional} {Training}.
\newblock In \emph{SMC'98 Conference Proceedings. 1998 IEEE International
  Conference on Systems, Man, and Cybernetics (Cat. No. 98CH36218)}, volume~2,
  pp.\  1968--1971, 1998.

\bibitem[Ghosh et~al.(2019)Ghosh, Sajjadi, Vergari, Black, and
  Sch{\"o}lkopf]{ghosh2019variational}
Partha Ghosh, Mehdi~SM Sajjadi, Antonio Vergari, Michael Black, and Bernhard
  Sch{\"o}lkopf.
\newblock From {Variational} to {Deterministic} {Autoencoders}.
\newblock In \emph{8th International Conference on Learning Representations,
  {ICLR}}, 2019.

\bibitem[Goodfellow et~al.(2014)Goodfellow, Pouget-Abadie, Mirza, Xu,
  Warde-Farley, Ozair, Courville, and
  Bengio]{Goodfellow2014GenerativeAdversarialNets}
Ian Goodfellow, Jean Pouget-Abadie, Mehdi Mirza, Bing Xu, David Warde-Farley,
  Sherjil Ozair, Aaron Courville, and Yoshua Bengio.
\newblock Generative {Adversarial} {Nets}.
\newblock In \emph{Advances in {Neural} {Information} {Processing} {Systems}
  27}, pp.\  2672--2680. 2014.

\bibitem[Han et~al.(2017)Han, Lu, Zhu, and Wu]{Han2017Alternating}
Tian Han, Yang Lu, Song~Chun Zhu, and Ying~Nian Wu.
\newblock Alternating {Back}-{Propagation} for {Generator} {Network}.
\newblock In \emph{Proceedings of the Thirty-First AAAI Conference on
  Artificial Intelligence}, pp.\  1976--1984, 2017.

\bibitem[Higgins et~al.(2017)Higgins, Matthey, Pal, Burgess, Glorot, Botvinick,
  Mohamed, and Lerchner]{higgins2017beta}
Irina Higgins, Loic Matthey, Arka Pal, Christopher Burgess, Xavier Glorot,
  Matthew Botvinick, Shakir Mohamed, and Alexander Lerchner.
\newblock Beta-{VAE}: {Learning} {Basic} {Visual} {Concepts} with a
  {Constrained} {Variational} {Framework}.
\newblock In \emph{5th International Conference on Learning Representations,
  {ICLR}}, 2017.

\bibitem[Hyv{\"a}rinen(2005)]{hyvarinen2005estimation}
Aapo Hyv{\"a}rinen.
\newblock Estimation of non-normalized statistical models by score matching.
\newblock \emph{Journal of Machine Learning Research}, 6\penalty0
  (Apr):\penalty0 695--709, 2005.

\bibitem[Ioffe \& Szegedy(2015)Ioffe and Szegedy]{ioffe2015batch}
Sergey Ioffe and Christian Szegedy.
\newblock Batch {Normalization}: {Accelerating} {Deep} {Network} {Training} by
  {Reducing} {Internal} {Covariate} {Shift}.
\newblock In \emph{Proceedings of the 32nd International Conference on Machine
  Learning, ICML}, pp.\  448–456, 2015.

\bibitem[Kamnitsas et~al.(2018)Kamnitsas, Castro, Folgoc, Walker, Tanno,
  Rueckert, Glocker, Criminisi, and Nori]{kamnitsas2018semi}
Konstantinos Kamnitsas, Daniel~C Castro, Loic~Le Folgoc, Ian Walker, Ryutaro
  Tanno, Daniel Rueckert, Ben Glocker, Antonio Criminisi, and Aditya Nori.
\newblock Semi-{Supervised} {Learning} via {Compact} {Latent} {Space}
  {Clustering}.
\newblock \emph{Proceedings of the 35th International Conference on Machine
  Learning}, 2018.

\bibitem[Karras et~al.(2020)Karras, Laine, Aittala, Hellsten, Lehtinen, and
  Aila]{karras2020analyzing}
Tero Karras, Samuli Laine, Miika Aittala, Janne Hellsten, Jaakko Lehtinen, and
  Timo Aila.
\newblock {Analyzing} and {Improving} the {Image} {Quality} of {StyleGAN}.
\newblock In \emph{Proceedings of the IEEE/CVF Conference on Computer Vision
  and Pattern Recognition}, pp.\  8110--8119, 2020.

\bibitem[Kingma \& Ba(2015)Kingma and Ba]{Kingma2017AdamMethodStochastic}
Diederik~P. Kingma and Jimmy Ba.
\newblock Adam: {A} {Method} for {Stochastic} {Optimization}.
\newblock In \emph{3rd International {Conference} on {Learning}
  {Representations}, ICLR}, 2015.

\bibitem[Kingma \& Welling(2014)Kingma and
  Welling]{Kingma2014AutoEncodingVariationalBayes}
Diederik~P. Kingma and Max Welling.
\newblock Auto-{Encoding} {Variational} {Bayes}.
\newblock \emph{2nd International Conference on Learning Representations,
  ICLR}, 2014.

\bibitem[Kingma \& Dhariwal(2018)Kingma and Dhariwal]{kingma2018glow}
Durk~P Kingma and Prafulla Dhariwal.
\newblock Glow: {Generative} {Flow} with {Invertible} 1x1 {Convolutions}.
\newblock In \emph{Advances in neural information processing systems}, pp.\
  10215--10224, 2018.

\bibitem[Krizhevsky et~al.(2009)Krizhevsky, Hinton,
  et~al.]{krizhevsky2009learning}
Alex Krizhevsky, Geoffrey Hinton, et~al.
\newblock Learning {Multiple} {Layers} of {Features} from {Tiny} {Images}.
\newblock \emph{Technical Report}, 2009.

\bibitem[LeCun et~al.(1998)LeCun, Bottou, Bengio, and
  Haffner]{lecun1998gradient}
Yann LeCun, L{\'e}on Bottou, Yoshua Bengio, and Patrick Haffner.
\newblock Gradient-{Based} {Learning} {Applied} to {Document} {Recognition}.
\newblock \emph{Proceedings of the IEEE}, 86\penalty0 (11):\penalty0
  2278--2324, 1998.

\bibitem[LeCun et~al.(2004)LeCun, Huang, and Bottou]{lecun2004learning}
Yann LeCun, Fu~Jie Huang, and Leon Bottou.
\newblock Learning {Methods} for {Generic} {Object} {Recognition} with
  {Invariance} to {Pose} and {Lighting}.
\newblock In \emph{Proceedings of the IEEE Computer Society Conference on
  Computer Vision and Pattern Recognition, CVPR}, volume~2, pp.\  II--104,
  2004.

\bibitem[Li et~al.(2019)Li, Liu, Liu, Zhao, and Liu]{li2019neural}
Naihan Li, Shujie Liu, Yanqing Liu, Sheng Zhao, and Ming Liu.
\newblock Neural {Speech} {Synthesis} with {Transformer} {Network}.
\newblock In \emph{Proceedings of the AAAI Conference on Artificial
  Intelligence}, volume~33, pp.\  6706--6713, 2019.

\bibitem[Lipton \& Tripathi(2017)Lipton and Tripathi]{lipton2017precise}
Zachary~C Lipton and Subarna Tripathi.
\newblock Precise {Recovery} of {Latent} {Vectors} from {Generative}
  {Adversarial} {Networks}.
\newblock In \emph{5th International Conference on Learning Representations,
  {ICLR}}, 2017.

\bibitem[Liu et~al.(2015)Liu, Luo, Wang, and Tang]{liu2015faceattributes}
Ziwei Liu, Ping Luo, Xiaogang Wang, and Xiaoou Tang.
\newblock Deep {Learning} {Face} {Attributes} in the {Wild}.
\newblock In \emph{Proceedings of International Conference on Computer Vision
  (ICCV)}, 2015.

\bibitem[Miyasawa(1961)]{miyasawa1961empirical}
Koichi Miyasawa.
\newblock An {Empirical} {Bayes} {Estimator} of the {Mean} of a {Normal}
  {Population}.
\newblock \emph{Bulletin of the International Statistical Institute},
  38\penalty0 (4):\penalty0 181--188, 1961.

\bibitem[Nane et~al.(1996)Nane, Nayar, and Murase]{nane1996columbia}
SA~Nane, SK~Nayar, and H~Murase.
\newblock Columbia {Object} {Image} {Library}: {COIL}-20.
\newblock \emph{Technical Report CUCS-005-96, Columbia University}, 1996.

\bibitem[Nijkamp et~al.(2019)Nijkamp, Hill, Zhu, and Wu]{nijkamp2019learning}
Erik Nijkamp, Mitch Hill, Song-Chun Zhu, and Ying~Nian Wu.
\newblock Learning {Non}-{Convergent} {Non}-{Persistent} {Short}-{Run} {MCMC}
  {Toward} {Energy}-{Based} {Model}.
\newblock In \emph{Neural Information Processing Systems (NeurIPS)}, pp.\
  5232--5242, 2019.

\bibitem[Nijkamp et~al.(2020)Nijkamp, Pang, Han, Zhou, Zhu, and
  Wu]{Nijkamp2020LearningMultilayerLatent}
Erik Nijkamp, Bo~Pang, Tian Han, Linqi Zhou, Song-Chun Zhu, and Ying~Nian Wu.
\newblock Learning {Multi-layer} {Latent} {Variable} {Model} via {Variational}
  {Optimization} of {Short} {Run} {MCMC} for {Approximate} {Inference}.
\newblock \emph{stat}, 1050:\penalty0 17, 2020.

\bibitem[Park et~al.(2019)Park, Florence, Straub, Newcombe, and
  Lovegrove]{Park2019DeepSDFLearningContinuous}
Jeong~Joon Park, Peter Florence, Julian Straub, Richard Newcombe, and Steven
  Lovegrove.
\newblock {DeepSDF}: {Learning} {Continuous} {Signed} {Distance} {Functions}
  for {Shape} {Representation}.
\newblock In \emph{Proceedings of the {IEEE} {Conference} on {Computer}
  {Vision} and {Pattern} {Recognition}}, pp.\  165--174, 2019.

\bibitem[Rezende \& Mohamed(2015)Rezende and
  Mohamed]{Rezende2015VariationalInferenceNormalizing}
Danilo~Jimenez Rezende and Shakir Mohamed.
\newblock Variational {Inference} with {Normalizing} {Flows}.
\newblock \emph{32nd International Conference on Machine Learning, ICML},
  2:\penalty0 1530--1538, 2015.

\bibitem[Robbins(1956)]{robbins1956empirical}
Herbert Robbins.
\newblock An {Empirical} {Bayes} {Approach} to {Statistics}.
\newblock In \emph{Proc. Third Berkely Symp.}, volume~1, pp.\  157--163, 1956.

\bibitem[Saremi \& Hyvarinen(2019)Saremi and Hyvarinen]{saremi2019neural}
Saeed Saremi and Aapo Hyvarinen.
\newblock Neural {Empirical} {Bayes}.
\newblock \emph{Journal of Machine Learning Research}, 20:\penalty0 1--23,
  2019.

\bibitem[Segler et~al.(2018)Segler, Preuss, and Waller]{segler2018planning}
Marwin~HS Segler, Mike Preuss, and Mark~P Waller.
\newblock Planning chemical syntheses with deep neural networks and symbolic
  {AI}.
\newblock \emph{Nature}, 555\penalty0 (7698):\penalty0 604--610, 2018.

\bibitem[Sitzmann et~al.(2020{\natexlab{a}})Sitzmann, Chan, Tucker, Snavely,
  and Wetzstein]{Sitzmann2020MetaSDFMetalearningSigned}
Vincent Sitzmann, Eric~R. Chan, Richard Tucker, Noah Snavely, and Gordon
  Wetzstein.
\newblock {MetaSDF}: {Meta}-learning {Signed} {Distance} {Functions}.
\newblock In \emph{eural Information Processing Systems (NeurIPS)},
  2020{\natexlab{a}}.

\bibitem[Sitzmann et~al.(2020{\natexlab{b}})Sitzmann, Martel, Bergman, Lindell,
  and Wetzstein]{Sitzmann2020ImplicitNeuralRepresentationsa}
Vincent Sitzmann, Julien N.~P. Martel, Alexander~W. Bergman, David~B. Lindell,
  and Gordon Wetzstein.
\newblock Implicit {Neural} {Representations} with {Periodic} {Activation}
  {Functions}.
\newblock In \emph{Neural Information Processing Systems (NeurIPS)},
  2020{\natexlab{b}}.

\bibitem[Song \& Ermon(2019)Song and Ermon]{song2019generative}
Yang Song and Stefano Ermon.
\newblock Generative {Modeling} by {Estimating} {Gradients} of the {Data}
  {Distribution}.
\newblock In \emph{Neural Information Processing Systems (NeurIPS)}, pp.\
  11918--11930. 2019.

\bibitem[Tancik et~al.(2020)Tancik, Srinivasan, Mildenhall, Fridovich-Keil,
  Raghavan, Singhal, Ramamoorthi, Barron, and Ng]{Tancik2020FourierFeaturesLet}
Matthew Tancik, Pratul~P. Srinivasan, Ben Mildenhall, Sara Fridovich-Keil,
  Nithin Raghavan, Utkarsh Singhal, Ravi Ramamoorthi, Jonathan~T. Barron, and
  Ren Ng.
\newblock Fourier {Features} {Let} {Networks} {Learn} {High} {Frequency}
  {Functions} in {Low} {Dimensional} {Domains}.
\newblock In \emph{Neural Information Processing Systems (NeurIPS)}, 2020.

\bibitem[Vahdat \& Kautz(2020)Vahdat and Kautz]{vahdat2020NVAE}
Arash Vahdat and Jan Kautz.
\newblock {NVAE}: A {Deep} {Hierarchical} {Variational} {Autoencoder}.
\newblock In \emph{Neural Information Processing Systems (NeurIPS)}, 2020.

\bibitem[Van Den~Oord et~al.(2016)Van Den~Oord, Kalchbrenner, and
  Kavukcuoglu]{VanDenOord2016PixelRecurrentNeural}
Aäron Van Den~Oord, Nal Kalchbrenner, and Koray Kavukcuoglu.
\newblock Pixel {Recurrent} {Neural} {Networks}.
\newblock In \emph{33rd {International} {Conference} on {Machine} {Learning},
  {ICML} 2016}, volume~4, pp.\  2611--2620, 2016.

\bibitem[Xiao et~al.(2017)Xiao, Rasul, and Vollgraf]{xiao2017fashion}
Han Xiao, Kashif Rasul, and Roland Vollgraf.
\newblock Fashion-{MNIST}: a {Novel} {Image} {Dataset} for {Benchmarking}
  {Machine} {Learning} {Algorithms}.
\newblock \emph{arXiv preprint arXiv:1708.07747}, 2017.

\bibitem[Yu et~al.(2015)Yu, Seff, Zhang, Song, Funkhouser, and
  Xiao]{yu2015lsun}
Fisher Yu, Ari Seff, Yinda Zhang, Shuran Song, Thomas Funkhouser, and Jianxiong
  Xiao.
\newblock {LSUN}: {Construction} of a {Large}-{Scale} {Image} {Dataset} using
  {Deep} {Learning} with {Humans} in the {Loop}.
\newblock \emph{arXiv preprint arXiv:1506.03365}, 2015.

\bibitem[Zadeh et~al.(2019)Zadeh, Lim, Liang, and
  Morency]{Zadeh2019VariationalAutoDecoder}
Amir Zadeh, Yao-Chong Lim, Paul~Pu Liang, and Louis-Philippe Morency.
\newblock Variational {Auto}-{Decoder}.
\newblock \emph{arXiv preprint arXiv:1903.00840}, 2019.

\bibitem[Zhao et~al.(2017)Zhao, Song, and Ermon]{zhao2017towards}
Shengjia Zhao, Jiaming Song, and Stefano Ermon.
\newblock Towards {Deeper} {Understanding} of {Variational} {Autoencoding}
  {Models}.
\newblock \emph{Proceedings of the 34th International Conference on Machine
  Learning}, 2017.

\bibitem[Zhou et~al.(2003)Zhou, Bousquet, Lal, Weston, and
  Sch{\"o}lkopf]{zhou2003learning}
Dengyong Zhou, Olivier Bousquet, Thomas Lal, Jason Weston, and Bernhard
  Sch{\"o}lkopf.
\newblock Learning with {Local} and {Global} {Consistency}.
\newblock \emph{Advances in neural information processing systems},
  16:\penalty0 321--328, 2003.

\bibitem[Zhu et~al.(2016)Zhu, Kr{\"a}henb{\"u}hl, Shechtman, and
  Efros]{zhu2016generative}
Jun-Yan Zhu, Philipp Kr{\"a}henb{\"u}hl, Eli Shechtman, and Alexei~A Efros.
\newblock Generative {Visual} {Manipulation} on the {Natural} {Image}
  {Manifold}.
\newblock In \emph{European Conference on Computer Vision}, pp.\  597--613,
  2016.

\end{thebibliography}
\bibliographystyle{iclr2021_conference}

\appendix

\section{Proof of Conditional Empirical Bayes \label{apx:empirical-bayes-proof}}

For latent variables $\mathbf{z} \sim p_\mathbf{z}$, a noisy observation of $\mathbf{z}$, $\mathbf{z}_0 = \mathbf{z} + \mathcal{N}(\mathbf{0},\bm{I}_d)$, and data point $\mathbf{x} \sim p_d$, we wish to find an estimator of $p(\mathbf{z}|\mathbf{x})$. To achieve this, we condition the Bayes least squares estimator on $\mathbf{x}$:
\begin{equation}
    \hat{\mathbf{z}}_\mathbf{x}(\mathbf{z}_0) = \int \mathbf{z} p(\mathbf{z}|\mathbf{z}_0,\mathbf{x}) d\mathbf{z} = \int \mathbf{z} \frac{p(\mathbf{z}_0, \mathbf{z} | \mathbf{x})}{p(\mathbf{z}_0|\mathbf{x})} d\mathbf{z}.
    \label{eqn:gon-bayes-estimator}
\end{equation}
Through the definition of the probabilistic chain rule and by marginalising out $\mathbf{z}$, we can define $p(\mathbf{z}_0|\mathbf{x}) = \int p(\mathbf{z}_0|\mathbf{z},\mathbf{x})p(\mathbf{z}|\mathbf{x}) d\mathbf{z}$ which can be simplified to $\int p(\mathbf{z}_0|\mathbf{z})p(\mathbf{z}|\mathbf{x}) d\mathbf{z}$ since $\mathbf{z}_0$ is dependent only on $\mathbf{z}$. Writing this out fully, we obtain:
\begin{equation}
    p(\mathbf{z}_0|\mathbf{x}) = \int \frac{1}{(2\pi)^{d/2}|\det(\bm{\Sigma})|^{1/2}}\exp \Big( -(\mathbf{z}_0-\mathbf{z})^T \bm{\Sigma}^{-1}(\mathbf{z}_0-\mathbf{z})/2 \Big) p(\mathbf{z}|\mathbf{x}) d\mathbf{z}.
\end{equation}
Differentiating with respect to $\mathbf{z}_0$ and multiplying both sides by $\bm{\Sigma}$ gives:
\begin{align}
    \bm{\Sigma} \nabla_{\mathbf{z}_0} p(\mathbf{z}_0 |\mathbf{x}) &= \int (\mathbf{z} - \mathbf{z}_0) p(\mathbf{z}_0|\mathbf{z}, \mathbf{x})p(\mathbf{z}|\mathbf{x}) d\mathbf{z} = \int (\mathbf{z}-\mathbf{z}_0) p(\mathbf{z}_0,\mathbf{z}|\mathbf{x}) d\mathbf{z} \\
    &= \int \mathbf{z}p(\mathbf{z}_0,\mathbf{z}|\mathbf{x}) d\mathbf{z} - \mathbf{z}_0 p(\mathbf{z}_0|\mathbf{x}).
\end{align}
After dividing both sides by $p(\mathbf{z}_0|\mathbf{x})$ and combining with Equation~\ref{eqn:gon-bayes-estimator} we get:
\begin{equation}
    \bm{\Sigma} \frac{\nabla_{\mathbf{z}_0} p(\mathbf{z}_0|\mathbf{x})}{p(\mathbf{z}_0|\mathbf{x})} = \int \mathbf{z} \frac{p(\mathbf{z}_0,\mathbf{z}|\mathbf{x})}{p(\mathbf{z}_0|\mathbf{x})} d\mathbf{z} - \mathbf{z}_0 = \hat{\mathbf{z}}_\mathbf{x}(\mathbf{z}_0) - \mathbf{z}_0.
\end{equation}
Finally, this can be rearranged to give the single step estimator of $\mathbf{z}$:
\begin{equation}
    \hat{\mathbf{z}}_\mathbf{x}(\mathbf{z}_0) = \mathbf{z}_0 + \bm{\Sigma}\frac{\nabla_{\mathbf{z}_0} p(\mathbf{z}_0|\mathbf{x})}{p(\mathbf{z}_0|\mathbf{x})} = \mathbf{z}_0 + \bm{\Sigma}\nabla_{\mathbf{z}_0} \log p(\mathbf{z}_0|\mathbf{x}).
\end{equation}

\section{Comparison with Autoencoders}
\label{apx:compare-autoencoders}

This section compares Gradient Origin Networks with autoencoders that use exactly the same architecture, as described in the paper. All experiments are performed with the CIFAR-10 dataset. After each epoch the model is fixed and the mean reconstruction loss is measured over the training set. 

\begin{figure}[H]

\begin{subfigure}{0.32\textwidth}
\begin{adjustbox}{width=\linewidth}
\begin{tikzpicture}
\begin{axis}[
    width=7cm,height=6cm,
    xlabel={Training Epoch},
    ylabel={Reconstruction Loss},
    xtick pos=left,
    ytick pos=left,
    enlarge x limits=false,
    every x tick/.style={color=black, thin},
    every y tick/.style={color=black, thin},
    tick align=outside,
    xlabel near ticks,
    ylabel near ticks,
    axis on top,
    legend style={draw=none},
    legend columns = 2
]
\addplot+[spectral2, mark options={fill=spectral2}, mark repeat=5] table [x=Epoch, y=AEmean, col sep=comma] {figures/gon-vs-ae/gon-vs-ae-4x4-nz=512,f=16,mean,std.csv};\addlegendentry{AE 4x4}
\addplot+[spectral3, mark options={fill=spectral3}, mark repeat=5] table [x=Epoch, y=GONmean, col sep=comma] {figures/gon-vs-ae/gon-vs-ae-4x4-nz=512,f=16,mean,std.csv};\addlegendentry{GON 4x4}

\addplot+[spectral10, mark options={fill=spectral10}, mark repeat=5] table [x=Epoch, y=AEmean, col sep=comma] {figures/gon-vs-ae/gon-vs-ae-6x6-nz=512,f=16,mean,std.csv};\addlegendentry{AE 6x6}
\addplot+[spectral4, mark options={fill=spectral4}, mark repeat=5] table [x=Epoch, y=GONmean, col sep=comma] {figures/gon-vs-ae/gon-vs-ae-6x6-nz=512,f=16,mean,std.csv};\addlegendentry{GON 6x6}

\addplot+[spectral5, mark options={fill=spectral5}, mark repeat=5] table [x=Epoch, y=AEmean, col sep=comma] {figures/gon-vs-ae/gon-vs-ae-8x8-nz=512,f=16,mean,std.csv};\addlegendentry{AE 8x8}
\addplot+[spectral8, mark options={fill=spectral8}, mark repeat=5] table [x=Epoch, y=GONmean, col sep=comma] {figures/gon-vs-ae/gon-vs-ae-8x8-nz=512,f=16,mean,std.csv};\addlegendentry{GON 8x8}

\addplot [name path=AE1upper,draw=none] table [x=Epoch,y expr=\thisrow{AEmean}+\thisrow{AEstd}, col sep=comma] {figures/gon-vs-ae/gon-vs-ae-4x4-nz=512,f=16,mean,std.csv};
\addplot [name path=AE1lower,draw=none] table [x=Epoch,y expr=\thisrow{AEmean}-\thisrow{AEstd}, col sep=comma] {figures/gon-vs-ae/gon-vs-ae-4x4-nz=512,f=16,mean,std.csv};
\addplot [fill=spectral2!20] fill between[of=AE1upper and AE1lower];

\addplot [name path=GON1upper,draw=none] table [x=Epoch,y expr=\thisrow{GONmean}+\thisrow{GONstd}, col sep=comma] {figures/gon-vs-ae/gon-vs-ae-4x4-nz=512,f=16,mean,std.csv};
\addplot [name path=GON1lower,draw=none] table [x=Epoch,y expr=\thisrow{GONmean}-\thisrow{GONstd}, col sep=comma] {figures/gon-vs-ae/gon-vs-ae-4x4-nz=512,f=16,mean,std.csv};
\addplot [fill=spectral3!20] fill between[of=GON1upper and GON1lower];

\addplot [name path=AE2upper,draw=none] table [x=Epoch,y expr=\thisrow{AEmean}+\thisrow{AEstd}, col sep=comma] {figures/gon-vs-ae/gon-vs-ae-6x6-nz=512,f=16,mean,std.csv};
\addplot [name path=AE2lower,draw=none] table [x=Epoch,y expr=\thisrow{AEmean}-\thisrow{AEstd}, col sep=comma] {figures/gon-vs-ae/gon-vs-ae-6x6-nz=512,f=16,mean,std.csv};
\addplot [fill=spectral10!20] fill between[of=AE2upper and AE2lower];

\addplot [name path=GON2upper,draw=none] table [x=Epoch,y expr=\thisrow{GONmean}+\thisrow{GONstd}, col sep=comma] {figures/gon-vs-ae/gon-vs-ae-6x6-nz=512,f=16,mean,std.csv};
\addplot [name path=GON2lower,draw=none] table [x=Epoch,y expr=\thisrow{GONmean}-\thisrow{GONstd}, col sep=comma] {figures/gon-vs-ae/gon-vs-ae-6x6-nz=512,f=16,mean,std.csv};
\addplot [fill=spectral4!20] fill between[of=GON2upper and GON2lower];

\addplot [name path=AE4upper,draw=none] table [x=Epoch,y expr=\thisrow{AEmean}+\thisrow{AEstd}, col sep=comma] {figures/gon-vs-ae/gon-vs-ae-8x8-nz=512,f=16,mean,std.csv};
\addplot [name path=AE4lower,draw=none] table [x=Epoch,y expr=\thisrow{AEmean}-\thisrow{AEstd}, col sep=comma] {figures/gon-vs-ae/gon-vs-ae-8x8-nz=512,f=16,mean,std.csv};
\addplot [fill=spectral5!20] fill between[of=AE4upper and AE4lower];

\addplot [name path=GON4upper,draw=none] table [x=Epoch,y expr=\thisrow{GONmean}+\thisrow{GONstd}, col sep=comma] {figures/gon-vs-ae/gon-vs-ae-8x8-nz=512,f=16,mean,std.csv};
\addplot [name path=GON4lower,draw=none] table [x=Epoch,y expr=\thisrow{GONmean}-\thisrow{GONstd}, col sep=comma] {figures/gon-vs-ae/gon-vs-ae-8x8-nz=512,f=16,mean,std.csv};
\addplot [fill=spectral8!20] fill between[of=GON4upper and GON4lower];

\end{axis}
\end{tikzpicture}
\end{adjustbox}
\caption{Encoding images to various spatial dimensions.}
\label{fig:gon-depth}
\end{subfigure}
\begin{subfigure}{0.32\textwidth}
\begin{adjustbox}{width=\linewidth}
\begin{tikzpicture}
\begin{axis}[
    width=7cm,height=6cm,
    xlabel={Training Epoch},
    ylabel={Reconstruction Loss},
    xtick pos=left,
    ytick pos=left,
    enlarge x limits=false,
    every x tick/.style={color=black, thin},
    every y tick/.style={color=black, thin},
    tick align=outside,
    xlabel near ticks,
    ylabel near ticks,
    axis on top,
    legend style={draw=none, fill=none, at={(0.025,1)},anchor=north west},
    legend columns=2,
]
\addplot+[spectral2, mark options={fill=spectral2}, mark repeat=5] table [x=Epoch, y=AEmean, col sep=comma] {figures/gon-vs-ae/gon-vs-ae-1x1-nz=512,f=16,mean,std.csv};\addlegendentry{AE $f\! =\! 16$}
\addplot+[spectral3, mark options={fill=spectral3}, mark repeat=5] table [x=Epoch, y=GONmean, col sep=comma] {figures/gon-vs-ae/gon-vs-ae-1x1-nz=512,f=16,mean,std.csv};\addlegendentry{GON $f\! =\! 16$}

\addplot+[spectral10, mark options={fill=spectral10}, mark repeat=5] table [x=Epoch, y=AEmean, col sep=comma] {figures/gon-vs-ae/gon-vs-ae-1x1-nz=512,f=64,mean,std.csv};\addlegendentry{AE $f\! =\! 64$}
\addplot+[spectral4, mark options={fill=spectral4}, mark repeat=5] table [x=Epoch, y=GONmean, col sep=comma] {figures/gon-vs-ae/gon-vs-ae-1x1-nz=512,f=64,mean,std.csv};\addlegendentry{GON $f\! =\! 64$}

\addplot+[spectral5, mark options={fill=spectral5}, mark repeat=5] table [x=Epoch, y=AEmean, col sep=comma] {figures/gon-vs-ae/gon-vs-ae-1x1-nz=512,f=128,mean,std.csv};\addlegendentry{AE $f\! =\! 128$}
\addplot+[spectral8, mark options={fill=spectral8}, mark repeat=5] table [x=Epoch, y=GONmean, col sep=comma] {figures/gon-vs-ae/gon-vs-ae-1x1-nz=512,f=128,mean,std.csv};\addlegendentry{GON $f\! =\! 128$}

\addplot [name path=AE1upper,draw=none] table [x=Epoch,y expr=\thisrow{AEmean}+\thisrow{AEstd}, col sep=comma] {figures/gon-vs-ae/gon-vs-ae-1x1-nz=512,f=16,mean,std.csv};
\addplot [name path=AE1lower,draw=none] table [x=Epoch,y expr=\thisrow{AEmean}-\thisrow{AEstd}, col sep=comma] {figures/gon-vs-ae/gon-vs-ae-1x1-nz=512,f=16,mean,std.csv};
\addplot [fill=spectral2!20] fill between[of=AE1upper and AE1lower];

\addplot [name path=GON1upper,draw=none] table [x=Epoch,y expr=\thisrow{GONmean}+\thisrow{GONstd}, col sep=comma] {figures/gon-vs-ae/gon-vs-ae-1x1-nz=512,f=16,mean,std.csv};
\addplot [name path=GON1lower,draw=none] table [x=Epoch,y expr=\thisrow{GONmean}-\thisrow{GONstd}, col sep=comma] {figures/gon-vs-ae/gon-vs-ae-1x1-nz=512,f=16,mean,std.csv};
\addplot [fill=spectral3!20] fill between[of=GON1upper and GON1lower];

\addplot [name path=AE2upper,draw=none] table [x=Epoch,y expr=\thisrow{AEmean}+\thisrow{AEstd}, col sep=comma] {figures/gon-vs-ae/gon-vs-ae-1x1-nz=512,f=64,mean,std.csv};
\addplot [name path=AE2lower,draw=none] table [x=Epoch,y expr=\thisrow{AEmean}-\thisrow{AEstd}, col sep=comma] {figures/gon-vs-ae/gon-vs-ae-1x1-nz=512,f=64,mean,std.csv};
\addplot [fill=spectral10!20] fill between[of=AE2upper and AE2lower];

\addplot [name path=GON2upper,draw=none] table [x=Epoch,y expr=\thisrow{GONmean}+\thisrow{GONstd}, col sep=comma] {figures/gon-vs-ae/gon-vs-ae-1x1-nz=512,f=64,mean,std.csv};
\addplot [name path=GON2lower,draw=none] table [x=Epoch,y expr=\thisrow{GONmean}-\thisrow{GONstd}, col sep=comma] {figures/gon-vs-ae/gon-vs-ae-1x1-nz=512,f=64,mean,std.csv};
\addplot [fill=spectral4!20] fill between[of=GON2upper and GON2lower];

\addplot [name path=AE4upper,draw=none] table [x=Epoch,y expr=\thisrow{AEmean}+\thisrow{AEstd}, col sep=comma] {figures/gon-vs-ae/gon-vs-ae-1x1-nz=512,f=128,mean,std.csv};
\addplot [name path=AE4lower,draw=none] table [x=Epoch,y expr=\thisrow{AEmean}-\thisrow{AEstd}, col sep=comma] {figures/gon-vs-ae/gon-vs-ae-1x1-nz=512,f=128,mean,std.csv};
\addplot [fill=spectral5!20] fill between[of=AE4upper and AE4lower];

\addplot [name path=GON4upper,draw=none] table [x=Epoch,y expr=\thisrow{GONmean}+\thisrow{GONstd}, col sep=comma] {figures/gon-vs-ae/gon-vs-ae-1x1-nz=512,f=128,mean,std.csv};
\addplot [name path=GON4lower,draw=none] table [x=Epoch,y expr=\thisrow{GONmean}-\thisrow{GONstd}, col sep=comma] {figures/gon-vs-ae/gon-vs-ae-1x1-nz=512,f=128,mean,std.csv};
\addplot [fill=spectral8!20] fill between[of=GON4upper and GON4lower];
\end{axis}
\end{tikzpicture}
\end{adjustbox}
\caption{Varying capacities by changing the number of convolution filters $f$.}
\label{fig:gon-capacity}
\end{subfigure}
\begin{subfigure}{0.32\textwidth}
\begin{adjustbox}{width=\linewidth}
\begin{tikzpicture}
\begin{axis}[
    width=7cm,height=6cm,
    xlabel={Training Epoch},
    ylabel={Reconstruction Loss},
    xtick pos=left,
    ytick pos=left,
    enlarge x limits=false,
    every x tick/.style={color=black, thin},
    every y tick/.style={color=black, thin},
    tick align=outside,
    xlabel near ticks,
    ylabel near ticks,
    axis on top,
    legend style={draw=none, fill=none, at={(0.03,1)},anchor=north west},
    legend columns=2
]
\addplot+[spectral2, mark options={fill=spectral2}, mark repeat=5] table [x=Epoch, y=AEmean, col sep=comma] {figures/gon-vs-ae/gon-vs-ae-1x1-nz=512,f=16,mean,std,update.csv};\addlegendentry{AE $k\! = \!512$}
\addplot+[spectral3, mark options={fill=spectral3}, mark repeat=5] table [x=Epoch, y=GONmean, col sep=comma] {figures/gon-vs-ae/gon-vs-ae-1x1-nz=512,f=16,mean,std,update.csv};\addlegendentry{GON $k\! = \!512$}

\addplot+[spectral10, mark options={fill=spectral10}, mark repeat=5] table [x=Epoch, y=AEmean, col sep=comma] {figures/gon-vs-ae/gon-vs-ae-1x1-nz=128,f=16,mean,std.csv};\addlegendentry{AE $k\! = \!128$}
\addplot+[spectral4, mark options={fill=spectral4}, mark repeat=5] table [x=Epoch, y=GON, col sep=comma] {figures/gon-vs-ae/gon-vs-ae-1x1-nz=128,f=16,mean.csv};\addlegendentry{GON $k\! = \!128$}

\addplot+[spectral5, mark options={fill=spectral5}, mark repeat=5] table [x=Epoch, y=AE, col sep=comma] {figures/gon-vs-ae/gon-vs-ae-1x1-nz=32,f=16,mean.csv};\addlegendentry{AE $k\! = \!32$}
\addplot+[spectral8, mark options={fill=spectral8}, mark repeat=5] table [x=Epoch, y=GON, col sep=comma] {figures/gon-vs-ae/gon-vs-ae-1x1-nz=32,f=16,mean.csv};\addlegendentry{GON $k\! = \!32$}

\addplot [name path=AE1upper,draw=none] table [x=Epoch,y expr=\thisrow{AEmean}+\thisrow{AEstd}, col sep=comma] {figures/gon-vs-ae/gon-vs-ae-1x1-nz=512,f=16,mean,std,update.csv};
\addplot [name path=AE1lower,draw=none] table [x=Epoch,y expr=\thisrow{AEmean}-\thisrow{AEstd}, col sep=comma] {figures/gon-vs-ae/gon-vs-ae-1x1-nz=512,f=16,mean,std,update.csv};
\addplot [fill=spectral2!20] fill between[of=AE1upper and AE1lower];

\addplot [name path=GON1upper,draw=none] table [x=Epoch,y expr=\thisrow{GONmean}+\thisrow{GONstd}, col sep=comma] {figures/gon-vs-ae/gon-vs-ae-1x1-nz=512,f=16,mean,std,update.csv};
\addplot [name path=GON1lower,draw=none] table [x=Epoch,y expr=\thisrow{GONmean}-\thisrow{GONstd}, col sep=comma] {figures/gon-vs-ae/gon-vs-ae-1x1-nz=512,f=16,mean,std,update.csv};
\addplot [fill=spectral3!20] fill between[of=GON1upper and GON1lower];

\addplot [name path=AE2upper,draw=none] table [x=Epoch,y expr=\thisrow{AEmean}+\thisrow{AEstd}, col sep=comma] {figures/gon-vs-ae/gon-vs-ae-1x1-nz=128,f=16,mean,std.csv};
\addplot [name path=AE2lower,draw=none] table [x=Epoch,y expr=\thisrow{AEmean}-\thisrow{AEstd}, col sep=comma] {figures/gon-vs-ae/gon-vs-ae-1x1-nz=128,f=16,mean,std.csv};
\addplot [fill=spectral10!20] fill between[of=AE2upper and AE2lower];

\addplot [name path=GON2upper,draw=none] table [x=Epoch,y expr=\thisrow{GONmean}+\thisrow{GONstd}, col sep=comma] {figures/gon-vs-ae/gon-vs-ae-1x1-nz=128,f=16,mean,std.csv};
\addplot [name path=GON2lower,draw=none] table [x=Epoch,y expr=\thisrow{GONmean}-\thisrow{GONstd}, col sep=comma] {figures/gon-vs-ae/gon-vs-ae-1x1-nz=128,f=16,mean,std.csv};
\addplot [fill=spectral4!20] fill between[of=GON2upper and GON2lower];

\addplot [name path=AE4upper,draw=none] table [x=Epoch,y expr=\thisrow{AEmean}+\thisrow{AEstd}, col sep=comma] {figures/gon-vs-ae/gon-vs-ae-1x1-nz=32,f=16,mean,std.csv};
\addplot [name path=AE4lower,draw=none] table [x=Epoch,y expr=\thisrow{AEmean}-\thisrow{AEstd}, col sep=comma] {figures/gon-vs-ae/gon-vs-ae-1x1-nz=32,f=16,mean,std.csv};
\addplot [fill=spectral5!20] fill between[of=AE4upper and AE4lower];

\addplot [name path=GON4upper,draw=none] table [x=Epoch,y expr=\thisrow{GONmean}+\thisrow{GONstd}, col sep=comma] {figures/gon-vs-ae/gon-vs-ae-1x1-nz=32,f=16,mean,std.csv};
\addplot [name path=GON4lower,draw=none] table [x=Epoch,y expr=\thisrow{GONmean}-\thisrow{GONstd}, col sep=comma] {figures/gon-vs-ae/gon-vs-ae-1x1-nz=32,f=16,mean,std.csv};
\addplot [fill=spectral8!20] fill between[of=GON4upper and GON4lower];
\end{axis}
\end{tikzpicture}
\end{adjustbox}
\caption{Encoding images to a variety of latent space sizes.}
\label{fig:gon-latents}
\end{subfigure}

\caption{Experiments comparing convolutional GONs with autoencoders on CIFAR-10, where the GON uses exactly same architecture as the AE, without the encoder. (a) At the limit autoencoders tend towards the identity function whereas GONs are unable to operate with no parameters. As the number of network parameters increases (b) and the latent size decreases (c), the performance lead of GONs over AEs decreases due to diminishing returns/bottlenecking.}
\end{figure}
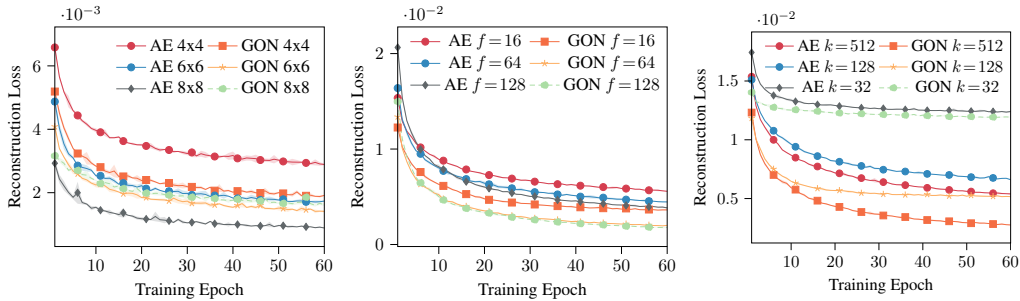

In Figure~\ref{fig:gon-depth} the depth of networks is altered by removing blocks thereby downscaling to various resolutions. We find that GONs outperform autoencoders until latents have a spatial size of 8x8 (where the equivalent GON now only has only 2 convolution layers). Considering the limit where neither model has any parameters, the latent space is the input data i.e. $\mathbf{z}=\mathbf{x}$. Substituting the definition of a GON (Equation \ref{eqn:gon}) this becomes $-\nabla_{\mathbf{z}_\mathbf{0}} \mathcal{L} = \mathbf{x}$ which simplifies to $\mathbf{0}=\mathbf{x}$ which is a contradiction. This is not a concern in normal practice, as evidenced by the results presented here.

Figure~\ref{fig:gon-capacity} explores the relationship between autoencoders and GONs when changing the number of convolution filters; GONs are found to outperform autoencoders in all cases. The discrepancy between loss curves decreases as the number of filters increases likely due to the diminishing returns of providing more capacity to the GON when its loss is significantly closer to 0.

A similar pattern is found when decreasing the latent space (Figure~\ref{fig:gon-latents}); in this case the latent space likely becomes the limiting factor. With larger latent spaces GONs significantly outperform autoencoders, however, when the latent bottleneck becomes more pronounced this lead lessens.

\section{Initialising $\mathbf{z}_0$ at the Origin}

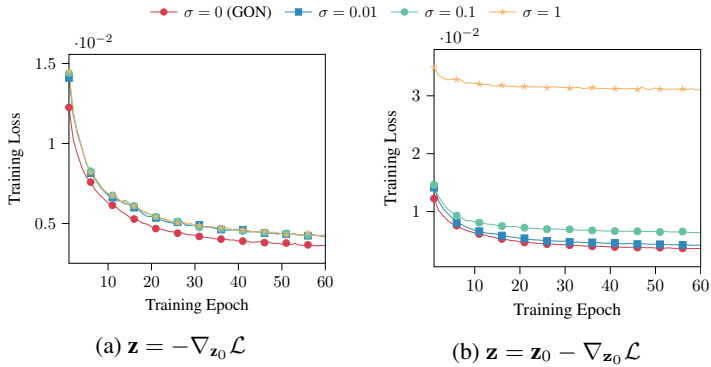
\begin{figure}[H]
\centering

\begin{adjustbox}{width=0.41\textwidth}
\begin{tikzpicture}
\matrix[
        matrix of nodes,
        anchor=south,
        draw=none,
        inner sep=0.2em,
    ] {
        \ref{plots:sigma0}& $\sigma=0$ (GON)&[4pt]
        \ref{plots:sigma0.01}& $\sigma=0.01$&[4pt]
        \ref{plots:sigma0.1}& $\sigma=0.1$&[4pt]
        \ref{plots:sigma1}& $\sigma=1$&[4pt]\\
    };
\end{tikzpicture}
\end{adjustbox}

\captionsetup[subfigure]{justification=centering}
\begin{subfigure}{0.32\textwidth}
\begin{adjustbox}{width=\linewidth}
\begin{tikzpicture}
\begin{axis}[
    width=7cm,height=6cm,
    xlabel={Training Epoch},
    ylabel={Training Loss},
    xtick pos=left,
    ytick pos=left,
    enlarge x limits=false,
    every x tick/.style={color=black, thin},
    every y tick/.style={color=black, thin},
    tick align=outside,
    xlabel near ticks,
    ylabel near ticks,
    axis on top,
    legend style={draw=none},
]
\addplot+[spectral2, mark options={fill=spectral2}, mark repeat=5] table [x=Epoch, y=GON, col sep=comma] {figures/gon-vs-ae/gon-vs-ae-1x1-nz=512,f=16,mean.csv};\addlegendentry{$\sigma=0$}\label{plots:sigma0}
\addplot+[spectral10, mark options={fill=spectral10}, mark repeat=5] table [x=Epoch, y index=1, col sep=comma] {figures/origin/origintest,cifar,nz=512,f=16,mean.csv};\addlegendentry{$\sigma=0.01$}\label{plots:sigma0.01}
\addplot+[spectral9, mark options={fill=spectral9}, mark repeat=5] table [x=Epoch, y index=2, col sep=comma] {figures/origin/origintest,cifar,nz=512,f=16,mean.csv};\addlegendentry{$\sigma=0.1$}\label{plots:sigma0.1}
\addplot+[spectral4, mark options={fill=spectral4}, mark repeat=5] table [x=Epoch, y index=3, col sep=comma] {figures/origin/origintest,cifar,nz=512,f=16,mean.csv};\addlegendentry{$\sigma=1$}\label{plots:sigma1}
\legend{}
\end{axis}
\end{tikzpicture}
\end{adjustbox}
\caption{$\textbf{z}=-\nabla_{\textbf{z}_0}\mathcal{L}$}
\label{fig:gon-origin-normal}
\end{subfigure}
\hspace*{1em}
\begin{subfigure}{0.32\textwidth}
\begin{adjustbox}{width=\linewidth}
\begin{tikzpicture}
\begin{axis}[
    width=7cm,height=6cm,
    xlabel={Training Epoch},
    ylabel={Training Loss},
    xtick pos=left,
    ytick pos=left,
    enlarge x limits=false,
    every x tick/.style={color=black, thin},
    every y tick/.style={color=black, thin},
    tick align=outside,
    xlabel near ticks,
    ylabel near ticks,
    axis on top,
    legend style={draw=none},
]
\addplot+[spectral2, mark options={fill=spectral2}, mark repeat=5] table [x=Epoch, y=GON, col sep=comma] {figures/gon-vs-ae/gon-vs-ae-1x1-nz=512,f=16,mean.csv};\addlegendentry{$\sigma=0$}\label{plots:sigma0}
\addplot+[spectral10, mark options={fill=spectral10}, mark repeat=5] table [x=Epoch, y index=1, col sep=comma] {figures/origin/origintest-graddescent,cifar,nz=512,f=16,mean.csv};\addlegendentry{$\sigma=0.01$}
\addplot+[spectral9, mark options={fill=spectral9}, mark repeat=5] table [x=Epoch, y index=2, col sep=comma] {figures/origin/origintest-graddescent,cifar,nz=512,f=16,mean.csv};\addlegendentry{$\sigma=0.1$}
\addplot+[spectral4, mark options={fill=spectral4}, mark repeat=5] table [x=Epoch, y index=3, col sep=comma] {figures/origin/origintest-graddescent,cifar,nz=512,f=16,mean.csv};\addlegendentry{$\sigma=1$}
\legend{}
\end{axis}
\end{tikzpicture}
\end{adjustbox}
\caption{$\mathbf{z}=\textbf{z}_0-\nabla_{\mathbf{z}_0}\mathcal{L}$}
\label{fig:gon-origin-descent}
\end{subfigure}
\caption{Training GONs with $\mathbf{z}_0$ sampled from a variety of normal distributions with different standard deviations $\sigma$, $\mathbf{z}_0\sim\mathcal{N}(\mathbf{0},\sigma^2\bm{I})$. Approach (a) directly uses the negative gradients as encodings while approach (b) performs one gradient descent style step initialised at $\mathbf{z}_0$.}
\label{fig:gon-origin}
\end{figure}

We evaluate different initialisations of $\mathbf{z}_0$ in Figure~\ref{fig:gon-origin} by sampling $\mathbf{z}_0 \sim \mathcal{N}(\mathbf{0}, \sigma^2\bm{I})$ for a variety of standard deviations $\sigma$. The proposed approach ($\sigma=0$) achieves the lowest reconstruction loss (Figure~\ref{fig:gon-origin-normal}); results for $\sigma>0$ are similar, suggesting that the latent space is adjusted so $\mathbf{z}_0$ simulates the origin. An alternative parameterisation of $\mathbf{z}$ is to use a use a single gradient descent style step $\mathbf{z}=\mathbf{z}_0 - \nabla_{\mathbf{z}_0}\mathcal{L}$ (Figure~\ref{fig:gon-origin-descent}), however, losses are higher than the proposed GON initialisation.

\section{Latent Space and Early Stopping}
\label{apx:early-stopping}

\begin{figure}[H]
\centering
\begin{minipage}{.32\textwidth}
  \vspace{1.2em}
  \centering
    \begin{adjustbox}{width=0.65\textwidth}
    \begin{tikzpicture}
    \begin{axis}[%
        width=2\textwidth,
        height=2\textwidth,
        axis line style={draw=none},
        tick style={draw=none},
        yticklabels={,,},
        xticklabels={,,},
        axis lines=middle,
        grid style={line width=1pt, draw=gray!10},
        major grid style={line width=0.5pt,draw=gray!50},
        axis line style={latex-latex},
        ]
        \addplot[
            scatter,%
            scatter/@pre marker code/.code={%
                \edef\temp{\noexpand\definecolor{mapped color}{rgb}{\pgfplotspointmeta}}%
                \temp
                \scope[draw=mapped color!100!black,fill=mapped color]%
            },%
            scatter/@post marker code/.code={%
                \endscope
            },%
            only marks,     
            mark=*,
            point meta={TeX code symbolic={%
                \edef\pgfplotspointmeta{\thisrow{red},\thisrow{green},\thisrow{blue}}%
            }},
        ] table {figures/embeddings/mnist,latent=2,hidden_features=128,layers=4,params=66851_2.dat};
    \end{axis}
    \end{tikzpicture}
    \end{adjustbox}
  \vspace{0.2em}
  \captionof{figure}{2D latent space samples of an implicit GON classifier trained on MNIST (class colours).\label{fig:latent}}
  \label{fig:test1}
\end{minipage}%
\hspace*{\fill}
\begin{minipage}{.32\textwidth}
  \centering
  \begin{adjustbox}{width=\textwidth}
    \begin{tikzpicture}
    \node[inner sep=0pt] (russell) at (0,0)
{\includegraphics[width=14em]{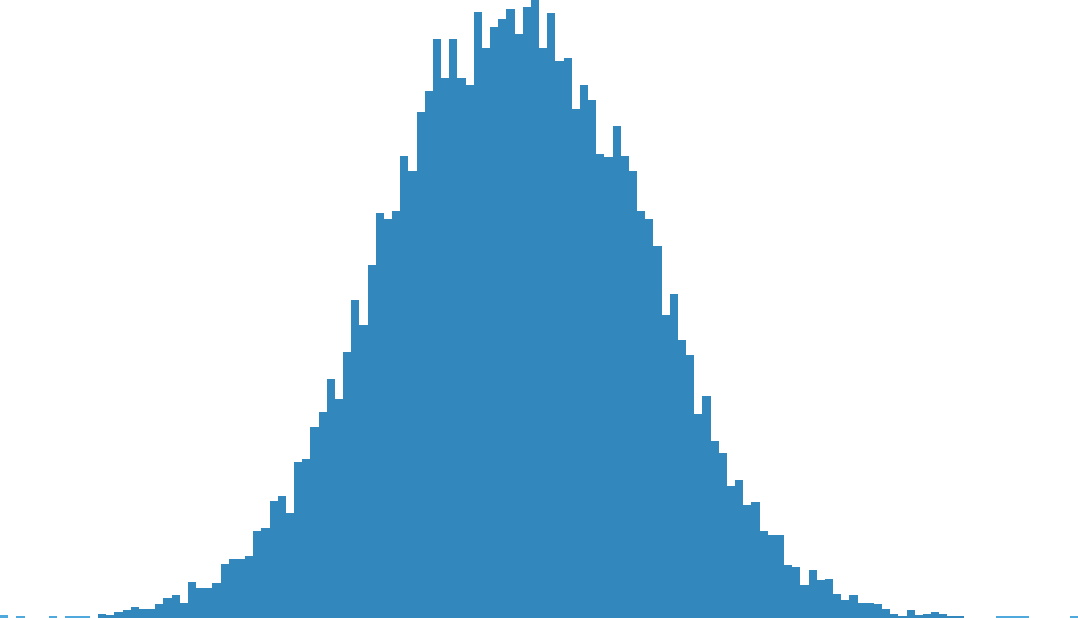}};
    \draw[->] (-2.47,-1.405)--(2.5,-1.405) node[right]{$x$};
    \draw[->] (-0.107,-1.405)--(-0.107,1.6) node[above]{$y$};  
    \end{tikzpicture}
  \end{adjustbox}
  \captionof{figure}{Histogram of latent gradients after 800 implicit GON steps with a SIREN.\label{fig:hist}}
  \label{fig:test}
\end{minipage}
\hspace*{\fill}
\begin{minipage}{.32\textwidth}
  \centering
  \begin{adjustbox}{width=\textwidth}
    \begin{tikzpicture}
    \node[inner sep=0pt] (russell) at (0,0)
{\includegraphics[width=14em]{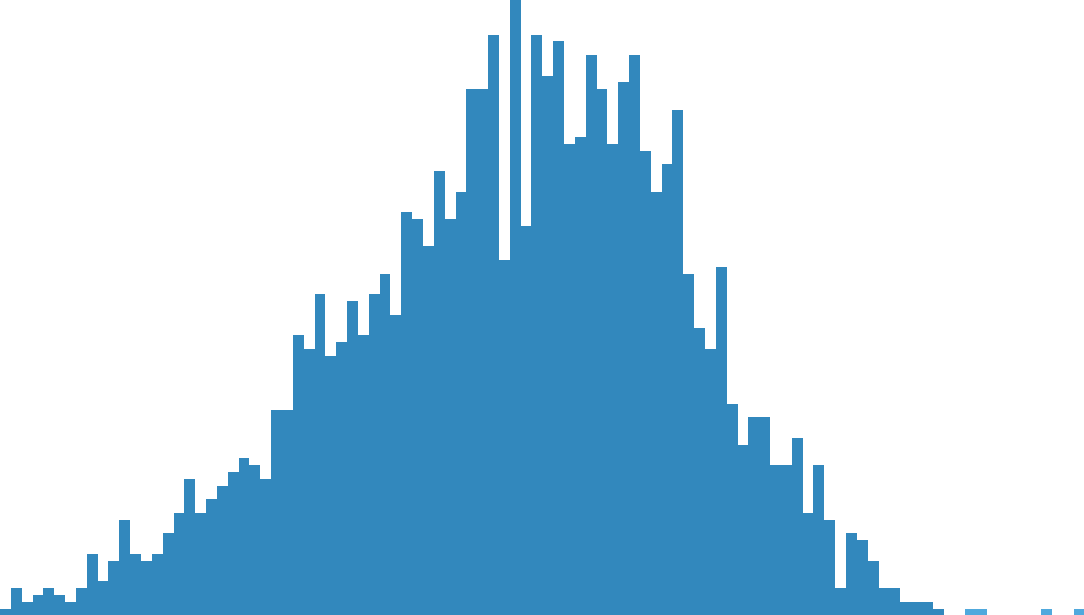}};
    \draw[->] (-2.47,-1.405)--(2.5,-1.405) node[right]{$x$};
    \draw[->] (-0.135,-1.405)--(-0.135,1.6) node[above]{$y$};  
    \end{tikzpicture}
  \end{adjustbox}
  \captionof{figure}{Histogram of traditional VAE latents after 800 steps.\label{fig:vae-hist}}
\end{minipage}
\end{figure}

Figure~\ref{fig:latent} shows a GON trained as a classifier where the latent space is squeezed into 2D for visualisation (Equation~\ref{eqn:gon-classify}) and Figure~\ref{fig:hist} shows that the negative gradients of a GON after 800 steps of training are approximately normally distributed; Figure~\ref{fig:vae-hist} shows that the latents of a typical convolutional VAE after 800 steps are significantly less normally distributed than GONs.

\begin{figure}[H]
    \centering
    \begin{subfigure}[t]{0.19\textwidth}
        \centering
        \includegraphics[width=\linewidth]{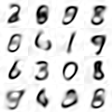}
        \caption{4,385 parameters}
    \end{subfigure}
    \hfill
    \begin{subfigure}[t]{0.19\textwidth}
        \centering
        \includegraphics[width=\linewidth]{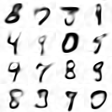}
        \caption{70k parameters}
    \end{subfigure}
    \hfill
    \begin{subfigure}[t]{0.19\textwidth}
       \centering
       \includegraphics[width=\linewidth]{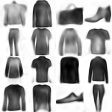}
        \caption{270k parameters}
    \end{subfigure}
    \hfill
    \begin{subfigure}[t]{0.19\textwidth}
       \centering
       \includegraphics[width=\linewidth]{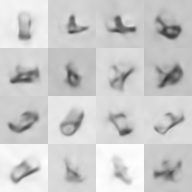}
        \caption{270k parameters}
    \end{subfigure}
    \hfill
    \begin{subfigure}[t]{0.19\textwidth}
       \centering
       \includegraphics[width=\linewidth]{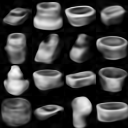}
        \caption{270k parameters}
    \end{subfigure}
\caption{GONs trained with early stopping can be sampled by approximating their latent space with a multivariate normal distribution. These images show samples from an implicit GON trained with early stopping.}

\label{fig:samples}
\end{figure}

To obtain new samples with implicit GONs, we can use early stopping as a simple alternative to variational approaches. These samples (Figure~\ref{fig:samples}) are diverse and capture the object shapes.

\section{GON Generalisation \label{apx:gon-generalisation}}

\begin{figure}[H]
\begin{subfigure}{0.32\textwidth}
\begin{adjustbox}{width=\linewidth}
\begin{tikzpicture}
\begin{axis}[
    width=7cm,height=6cm,
    xlabel={Training Epoch},
    ylabel={Reconstruction Loss},
    xtick pos=left,
    ytick pos=left,
    enlarge x limits=false,
    every x tick/.style={color=black, thin},
    every y tick/.style={color=black, thin},
    tick align=outside,
    xlabel near ticks,
    ylabel near ticks,
    axis on top,
    legend style={draw=none},
    ymax=0.005, ymin=0
]
\addplot+[spectral2, mark options={fill=spectral2}, mark repeat=50] table [x=Epoch, y={GON Train}, col sep=comma] {figures/long_plots/mnist_long.csv};\addlegendentry{GON Train}
\addplot+[spectral10, mark options={fill=spectral10}, mark repeat=50] table [x=Epoch, y={GON Test}, col sep=comma] {figures/long_plots/mnist_long.csv};\addlegendentry{GON Test}
\addplot+[spectral4, mark options={fill=spectral4}, mark repeat=50] table [x=Epoch, y={AE Train}, col sep=comma] {figures/long_plots/mnist_long.csv};\addlegendentry{AE Train}
\addplot+[spectral8, mark options={fill=spectral8}, mark repeat=50] table [x=Epoch, y={AE Test}, col sep=comma] {figures/long_plots/mnist_long.csv};\addlegendentry{AE Test}
\end{axis}
\end{tikzpicture}
\end{adjustbox}
\caption{MNIST}
\end{subfigure}
\begin{subfigure}{0.32\textwidth}
\begin{adjustbox}{width=\linewidth}
\begin{tikzpicture}
\begin{axis}[
    width=7cm,height=6cm,
    xlabel={Training Epoch},
    ylabel={Reconstruction Loss},
    xtick pos=left,
    ytick pos=left,
    enlarge x limits=false,
    every x tick/.style={color=black, thin},
    every y tick/.style={color=black, thin},
    tick align=outside,
    xlabel near ticks,
    ylabel near ticks,
    axis on top,
    legend style={draw=none},
    ymax=0.005, ymin=0
]
\addplot+[spectral2, mark options={fill=spectral2}, mark repeat=50] table [x=Epoch, y={GON Train}, col sep=comma] {figures/long_plots/fashion_long.csv};\addlegendentry{GON Train}
\addplot+[spectral10, mark options={fill=spectral10}, mark repeat=50] table [x=Epoch, y={GON Test}, col sep=comma] {figures/long_plots/fashion_long.csv};\addlegendentry{GON Test}
\addplot+[spectral4, mark options={fill=spectral4}, mark repeat=50] table [x=Epoch, y={AE Train}, col sep=comma] {figures/long_plots/fashion_long.csv};\addlegendentry{AE Train}
\addplot+[spectral8, mark options={fill=spectral8}, mark repeat=50] table [x=Epoch, y={AE Test}, col sep=comma] {figures/long_plots/fashion_long.csv};\addlegendentry{AE Test}
\end{axis}
\end{tikzpicture}
\end{adjustbox}
\caption{Fashion-MNIST}
\end{subfigure}
\begin{subfigure}{0.32\textwidth}
\begin{adjustbox}{width=\linewidth}
\begin{tikzpicture}
\begin{axis}[
    width=7cm,height=6cm,
    xlabel={Training Epoch},
    ylabel={Reconstruction Loss},
    xtick pos=left,
    ytick pos=left,
    enlarge x limits=false,
    every x tick/.style={color=black, thin},
    every y tick/.style={color=black, thin},
    tick align=outside,
    xlabel near ticks,
    ylabel near ticks,
    axis on top,
    legend style={draw=none},
    ymax=0.002, ymin=0
]
\addplot+[spectral2, mark options={fill=spectral2}, mark repeat=50] table [x=Epoch, y={GON Train}, col sep=comma] {figures/long_plots/norb_long.csv};\addlegendentry{GON Train}
\addplot+[spectral10, mark options={fill=spectral10}, mark repeat=50] table [x=Epoch, y={GON Test}, col sep=comma] {figures/long_plots/norb_long.csv};\addlegendentry{GON Test}
\addplot+[spectral4, mark options={fill=spectral4}, mark repeat=50] table [x=Epoch, y={AE Train}, col sep=comma] {figures/long_plots/norb_long.csv};\addlegendentry{AE Train}
\addplot+[spectral8, mark options={fill=spectral8}, mark repeat=50] table [x=Epoch, y={AE Test}, col sep=comma] {figures/long_plots/norb_long.csv};\addlegendentry{AE Test}
\end{axis}
\end{tikzpicture}
\end{adjustbox}
\caption{Small NORB}
\end{subfigure}

\caption{The discrepancy between training and test reconstruction losses when using a GON is smaller than equivalent autoencoders over a variety of datasets.}
\label{fig:gon-generalisation-datasets}
\end{figure}
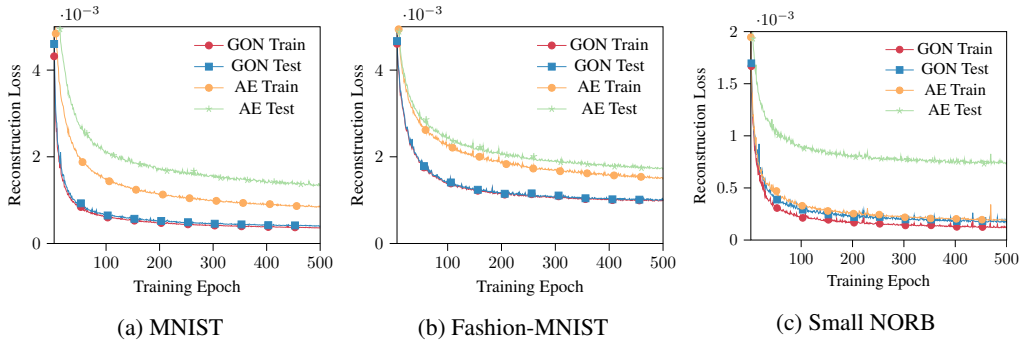

In Figure~\ref{fig:gon-generalisation-datasets} the training and test losses for GONs and autoencoders are plotted for a variety of datasets. In all cases, GONs generalise better than their equivalent autoencoders while achieving lower losses with fewer parameters.

\section{Availability}
Source code for the convolutional GON, variational GON, and implicit GON is available under the MIT license on GitHub at: \url{https://github.com/cwkx/GON}. This implementation uses PyTorch and all reported experiments use a Nvidia RTX 2080 Ti GPU.

\newpage
\section{Super-resolution}
\begin{figure}[H]
\centering
\begin{subfigure}{0.45\linewidth}
\includegraphics[width=\linewidth]{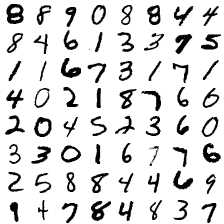}
\caption{Original data 28x28}
\end{subfigure}
\begin{subfigure}{0.45\linewidth}
\includegraphics[width=\linewidth]{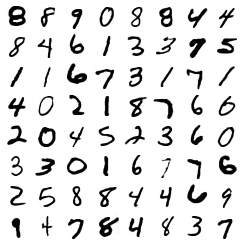}
\caption{Implicit GON at 28x28}
\end{subfigure}
\begin{subfigure}{0.9\linewidth}
\includegraphics[width=\linewidth]{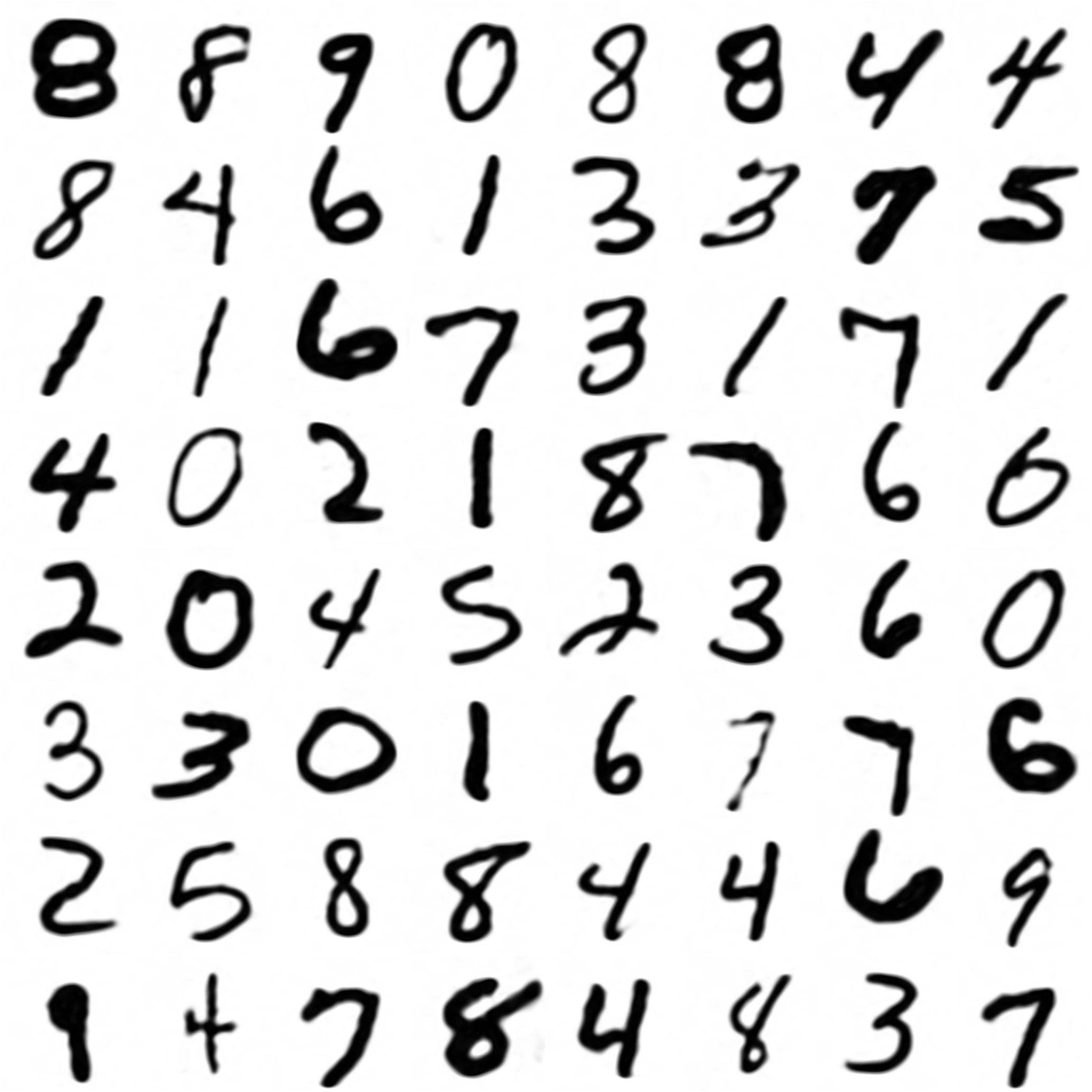}
\end{subfigure}
\caption{Super-sampling 28x28 MNIST test data at 256x256 coordinates using an implicit GON.}
\label{fig:gon-supermnist}
\end{figure}

Additional super-sampled images obtained by training an implicit GON on 28x28 MNIST data then reconstructing test data at a resolution of 256x256 are shown in Figure~\ref{fig:gon-supermnist}.

\end{document}